\documentclass[sn-mathphys]{sn-jnl}

\jyear{2025}%

\theoremstyle{thmstyleone}%
\theoremstyle{thmstyletwo}%

\theoremstyle{thmstylethree}%
\usepackage{graphicx}
\usepackage{url}
\usepackage{multirow}
\usepackage[table,xcdraw]{xcolor}
\usepackage{arydshln}
\usepackage{xcolor}
\usepackage{comment}
\definecolor{mydeepgreen}{rgb}{0,0.5,0}  

\begin{document}

\title[MindDiffuser]{Versatile Framework   with Semantic and Structural guidance for  Image Reconstruction from Brain Activity}

\author[1,2]{\fnm{Yizhuo} \sur{Lu}}
\equalcont{These authors contributed equally to this work.}

\author[1]{\fnm{Changde} \sur{Du}}
\equalcont{These authors contributed equally to this work.}

\author[1,3]{\fnm{Qiongyi} \sur{Zhou}}

\author[1,2]{\fnm{Liuyun} \sur{Jiang}}

\author[1,2,3]{\fnm{Huiguang} \sur{He}\textsuperscript{*}}

\affil[1]{\orgdiv{State Key Laboratory of Brain Cognition and Brain-inspired Intelligence Technology}, \orgname{Institute of Automation, Chinese Academy of Sciences}, \orgaddress{\city{Beijing} \postcode{100190},  \country{China}}}

\affil[2]{\orgdiv{School of Future Technology}, \orgname{University of Chinese Academy of Sciences}, \orgaddress{\city{Beijing} \postcode{100049},  \country{China}}}

\affil[3]{\orgdiv{School of Artificial Intelligence}, \orgname{University of Chinese Academy of Sciences}, \orgaddress{\city{Beijing} \postcode{100049},  \country{China}}}
\renewcommand{\thefootnote}{} 
\footnotetext{* Corresponding author. Email: huiguang.he@ia.ac.cn}
\renewcommand{\thefootnote}{\arabic{footnote}} 

\abstract{Reconstructing visual stimuli from brain recordings has been a meaningful and challenging task in brain decoding. Especially, the achievement of precise and controllable image reconstruction bears great significance in propelling the progress and utilization of brain-computer interfaces.  Recent methods, leveraging advances in the power of text-to-image generation models, have reconstructed images that closely approximate complex natural stimuli in terms of semantics (e.g., concepts and objects). However, they struggle to maintain consistency with the original stimuli in fine-grained structural information (e.g., position, orientation and size), which undermines both the controllability and interpretability of the models.  To address the aforementioned issues, we propose a two-stage image reconstruction framework, termed MindDiffuser. In Stage 1, Contrastive Language-Image Pretraining (CLIP) text embeddings decoded from brain responses are input into Stable Diffusion, generating a preliminary image containing semantic information. In Stage 2, we use decoded shallow CLIP visual features as supervisory signals, iteratively refining the feature vectors from Stage 1 via backpropagation to align structural information.  We conducted extensive experiments on brain response datasets across three modalities (fMRI, EEG, MEG) elicited by visual stimuli, demonstrating that our framework significantly enhances the performance of previous state-of-the-art models, highlighting the effectiveness and versatility of our approach. Spatial and temporal visualization results further support the neurobiological plausibility of our framework, providing guidance for future neural decoding efforts across different brain signal modalities.}




\maketitle
\section{Introduction}

\begin{figure}[htbp!]
	\centering
	\includegraphics[width=0.8\linewidth]{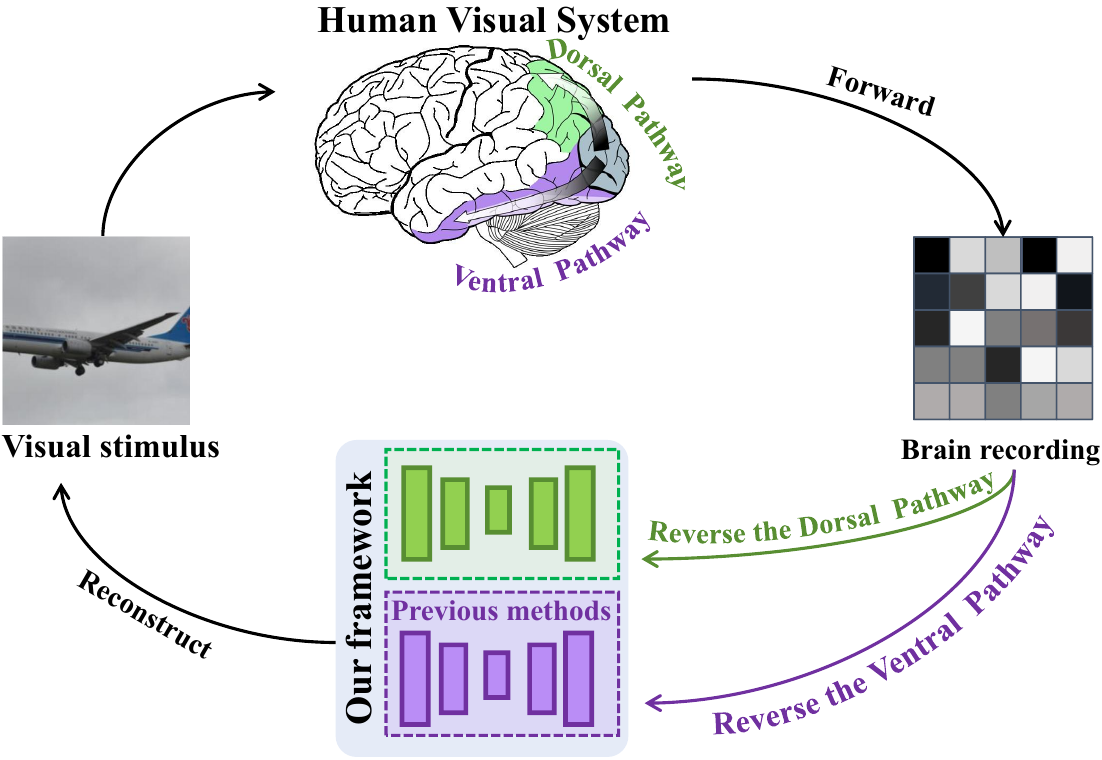}
	\caption{An illustration of the \textbf{forward} and \textbf{reverse} pathways in visual processing systems. Forward visual processing in the human brain involves the ventral stream, which processes semantic-level information like objects and concepts, and the dorsal stream, which processes structural-level spatial information such as position and orientation. Our framework reverses these two pathways to decode both semantic and structural features, enabling accurate reconstruction of stimulus images.}
	\label{fig:motivation}
\end{figure}

The human visual system excels at perceiving and interpreting complex visual stimuli with remarkable efficiency and robustness, far exceeding current AI models. Understanding the neural mechanisms behind these processes is crucial for unraveling brain function \cite{rakhimberdina2021natural, zhou2022exploring}, despite the challenges. A central focus of this research is neural decoding \cite{10047967, zhou2025interpretable}, which seeks to link visual stimuli with brain recordings. Neural decoding techniques encompass classification, identification, and reconstruction  tasks, with this study emphasizing reconstruction-the most challenging among them.

Early image reconstruction studies predominantly employed end-to-end paradigms for training \cite{du2020structured, du2018reconstructing, 10.3389/fncom.2019.00021, beliy2019voxels}.  However, due to the scarcity of paired image-stimulus and brain-response data in this field, these paradigms often yielded blurry reconstructions that lacked precise semantic content. Recent advancements in large-scale text-to-image (T2I) models, such as BigGAN \cite{donahue2019large} and Stable Diffusion \cite{rombach2022high}, have led to significant progress in the field. Trained on extensive image-text pair datasets, these models have acquired rich priors that enable them to generate highly realistic images based on textual prompts. This rapid development has garnered considerable attention from researchers in neuroscience.  Many studies first map fMRI data to the corresponding multimodal representations of stimulus images, and then input these representations into pre-trained T2I models. By leveraging the T2I model's powerful visual priors, they have successfully reconstructed semantically coherent natural images \cite{lin2022mind, ozcelik2023natural, takagi2023high, gu2022decoding, scotti2024reconstructing}.  Although this paradigm offers advantages such as lower training difficulty and reconstructions with clear semantic content, it also has notable drawbacks. Additionally, due to the lack of additional constraints on the T2I model, the reconstructed images fail to align with the original images in terms of structural information, including size, position, and shape.

To address this issue, we draw inspiration from the visual cortex processing mechanisms of the brain. As shown in Fig. \ref{fig:motivation}, in the human brain, forward visual information processing primarily relies on two pathways: the ventral stream and the dorsal stream. The ventral stream is responsible for processing semantic-level information such as concepts and objects, while the dorsal stream handles structural-level spatial information, including position and orientation. This collaboration enables humans to comprehend both \textbf{what} is present in an image and \textbf{where} it is located \cite{vaziri2017goal, zachariou2014ventral}.  It is evident that prior work within this framework has only reversed the ventral stream, neglecting the structural-level spatial information processed by the dorsal stream, which has led to suboptimal reconstruction outcomes. In this work, we propose MindDiffuser, a two-stage image reconstruction framework that reverses both the ventral and dorsal streams. In Stage 1, CLIP text embeddings decoded from brain responses are input into Stable Diffusion, generating a preliminary image that captures the semantic content, thus providing an initial understanding of ``\textbf{what} is in the image." In Stage 2, we use decoded shallow CLIP visual features as supervisory signals, iteratively refining the feature vectors from Stage 1 via backpropagation to align structural information, thereby constraining ``\textbf{where} the objects are located in the image."

Note that there is a previous conference version\footnote{Code are available at: https://github.com/ReedOnePeck/MindDiffuser} of this work \cite{lu2023minddiffuser}. In contrast to previous work, we extend our method from fMRI to EEG and MEG data, validating our framework on ten top models from various modalities and architectures. This version also includes extensive experimental and temporal/spatial interpretability analyses. Our contributions are summarized as follows:

\textbf{(1) Controllable framework:} Inspired by the visual cortex's information processing mechanisms, we propose a two-stage framework (Fig. \ref{fig:overview2}) that reverses the ventral and dorsal pathways, guided by semantic and structural information, addressing the limitation of previous methods that fail to incorporate both aspects.

\textbf{(2) Versatile performance:} We conducted extensive experiments on datasets from different modalities (fMRI, EEG and MEG) and reconstruction methods based on various generative models (GAN, Diffusion). The results demonstrate that our framework can be integrated into existing models to significantly enhance their low-level structural metrics, with minimal impact on high-level semantic metrics (Table \ref{tab:result2}).

\textbf{(3) Interpretability:} Considering the characteristics of different modality data, we conducted detailed spatial and temporal visualization analyses, revealing the interpretability and rationality of the framework in relation to the experimental results (Fig. \ref{fig:visualcortexsub1}).

\section{Related Work}

\subsection{Generative models for neural decoding}
Early image generation models primarily focused on VAE \cite{kingma2013auto} and GAN \cite{goodfellow2020generative}, which are lightweight but difficult to scale up. Recently, diffusion models \cite{wijmans1995solution, ho2020denoising} have rapidly emerged as a new generative paradigm. In these models, Gaussian noise is added to an image during the forward diffusion process until collapse, followed by a reverse denoising process to generate the image. With strong multimodal representations and large-scale text-image datasets, DALLE-2 \cite{ramesh2022hierarchical} and Stable Diffusion \cite{rombach2022high} have emerged as leading diffusion models. Additionally, models like T2I-Adapter \cite{mou2024t2i} and Versatile Diffusion \cite{xu2023versatile} have further enhanced generation quality by incorporating more refined control. Neuroscientists have also leveraged their powerful generative capabilities for image reconstruction tasks, yielding promising results.

\subsection{Image reconstruction from human brain}
Previous image reconstruction methods employed linear regression models to map fMRI data to \textbf{manually defined features}, resulting in blurry outputs and heavy reliance on manual feature selection \cite{kay2008naselaris,naselaris2009bayesian,fujiwara2013modular}. With the rise of deep learning,  Beliy et al. \cite{beliy2019voxels} and Gaziv et al. \cite{gaziv2022self} used semi-supervised learning \cite{chapelle2009semi} to train an Encoder-Decoder model for image reconstruction, addressing the issue of limited stimulus-fMRI pairs. Du et al. \cite{du2018reconstructing} introduced a multi-view reconstruction model that captures the statistical correlation between fMRI signals and stimuli. Despite improving performance, this \textbf{end-to-end paradigm} still struggles with the lack of clear semantic information in the reconstruction results, making them difficult to interpret. Recently, \textbf{pre-trained generative models} have increasingly been applied to image reconstruction tasks. Chen et al. \cite{chen2022seeing} and Takagi et al. \cite{takagi2023high} mapped fMRI data to the latent space of Stable Diffusion, while Ozcelik et al. \cite{ozcelik2022reconstruction} and Gu et al. \cite{gu2022decoding} used IC-GAN \cite{casanova2021instance} for image reconstruction. Although these methods reconstruct clear semantic information, \textbf{they overlook structural details such as position and size.}

\subsection{Multi-stage modeling in image reconstruction}
\label{section_multi_stage}
Recent works have sought to address the above issues. Xia et al. \cite{xia2024dream} explicitly decoded structural information such as color and depth from fMRI by reverse-engineering the Parvocellular and Magnocellular pathways, though the interpretability of their approach was not validated. Shen et al. \cite{shen2019deep} used VGG19 \cite{simonyan2014very} features as supervisory signals to iteratively optimize the latent space of a GAN for aligning structural features. However, due to the limited dataset size, the results lacked semantic information. Similarly, Kneeland et al. \cite{kneeland2023second} optimized the latent space of a generative model by incorporating an external neural encoding model, while Xie et al. \cite{xie2024brainram} used Retrieval-Augmented Generation (RAG) techniques to retrieve better image embeddings from historical training data for reconstruction. These methods primarily improve high-level semantic metrics, with little to no significant enhancement in low-level structural metrics. \textbf{The main goal of this work is to improve the alignment of structural information in the reconstruction results.}

\section{Methods}
\begin{figure*}[htbp!]
	\centering
	\includegraphics[width=0.95\linewidth]{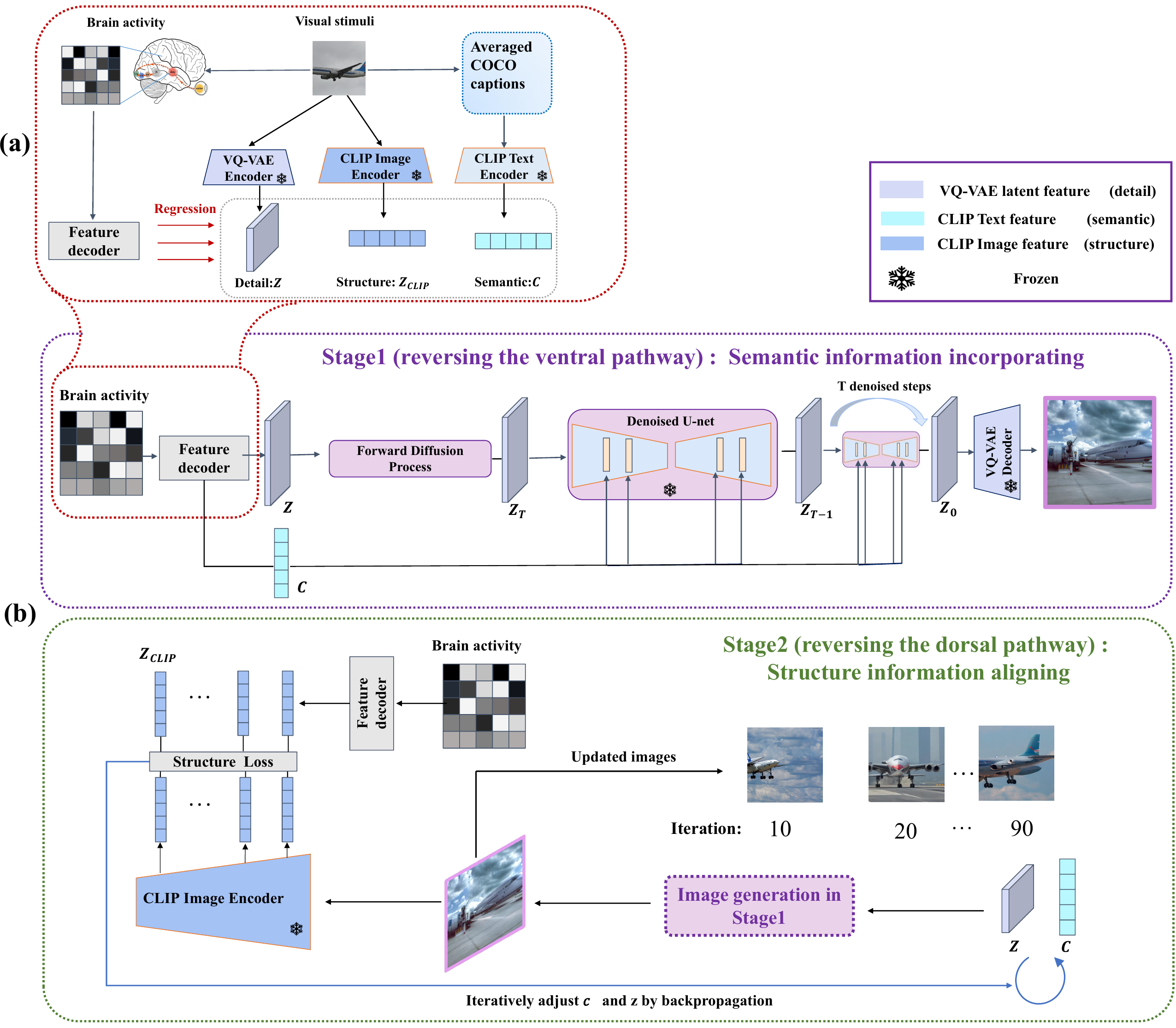}
	\caption{Schematic diagram of MindDiffuser. A set of brain decoders are trained to map brain signals to image features. (b) The two-stage image reconstruction process. In Stage 1, an initial reconstructed image is generated using the decoded CLIP text feature $c$ and VQ-VAE latent feature $z$. In Stage 2, the decoded CLIP image feature is used as a constraint to iteratively adjust $c$ and $z$ until the final reconstruction result matches the original image in terms of both semantic and structure.}
	\label{fig:overview2}
\end{figure*}
\subsection{Overview}

In this section, we introduce {\bfseries MindDiffuser}, a novel two-stage framework for controlled image reconstruction, as illustrated in Fig. \ref{fig:overview2}. Specifically, Stage 1 focuses on reversing the ventral visual pathway. In this stage, brain signals are decoded into CLIP text embeddings $c$ and visual features $z$ within the VQ-VAE latent space. This enables the initial reconstructed images generated by Stable Diffusion  to contain semantic information and coarse-grained content, thereby interpreting {\bfseries ``What is contained in the image ?''} Subsequently, Stage 2 targets reversing the dorsal visual pathway. Here, brain signals are transformed into shallow CLIP visual features, which are then used to iteratively refine $c$ and $z$ from Stage 1 through back-propagation. This iterative optimization allows the reconstructed images to closely align with the ground truth in the structural embedding space, enabling guidance over fine-grained structural information and answering the question: {\bfseries ``Where are the objects in the image ?"}

\subsection{Training}
Let $Y \in \mathbb{R}^{N \times 512 \times 512 \times 3}$ and $X \in \mathbb{R}^{N \times D_{voxel}}$ represent the visual stimuli and their corresponding fMRI activity patterns in the training set, respectively. Here, $N$ denotes the number of samples in the training set, and $D_{voxel}$ represents the number of voxels.
\subsubsection{Feature extracting}
We extracted the CLIP text branch features $c \in \mathbb{R}^{1 \times 15 \times 768}$ from Stable Diffusion to represent the semantic feature of the stimuli.  	In addition, we extracted features $z \in \mathbb{R}^{1 \times 4 \times 64 \times 64}$ from the VQ-VAE latent space of the images. These features were used as the initial noise input to Stable Diffusion, enabling the injection of coarse-grained content into the reconstruction process.

Researches by Zhou et al. \cite{zhou2022exploring} and Wang et al. \cite{Wang2022.09.27.508760} have demonstrated a hierarchical correspondence between CLIP features and the human visual cortex. Specifically, shallow layers of CLIP are associated with primary brain regions responsible for low-level structural processing, while deeper layers correspond to higher-order regions involved in semantic processing. Leveraging this insight, we extracted shallow features from the visual branch of CLIP as structural constraints, denoted as $Z_{CLIP}^i \in \mathbb{R}^{1 \times 38400}, i \in \{2,4,6,8,10,12\}$, derived from the linear layers of the 2nd, 4th, 6th, 8th, 10th, and 12th attention modules.

\subsubsection{Feature decoding}
To decode fMRI signals into the three aforementioned features, we employed simple linear regression models, enabling enhanced interpretability in subsequent analyses. As illustrated in Fig.  \ref{fig:overview2}(a), we trained the following models: $f_c: X \mapsto c$, $f_z: X \mapsto z$, and $f_{CLIPi}: X \mapsto Z_{CLIP}^i$. The ground-truth feature is $y$, and the predicted feature is $\hat{y}$. The loss function for these models is expressed as:

\begin{equation}\label{equ111}
	L = \|y - \hat{y}\|_2 + \alpha \|W\|_2,
\end{equation}

where $W$ represents the model weights, and $\alpha$ denotes the regularization factor.

Once these linear models were trained, the weights at each voxel represent the importance of it in decoding a specific feature. By projecting these model weights onto the cortical surface, we can directly visualize importance maps for decoding semantic and structural features within the visual cortex.

To handle multi-channel EEG and MEG data, we adapt the widely used channel-wise attention, along with the Temporal and Spatial Convolution module \cite{li2024visual, song2023decoding}. The features are then flattened and integrated using an MLP to project them to the desired dimensions, as detailed in Table \ref{tab:EEG}.

\begin{table}[htbp!]
	\caption{Layer-wise parameters for the EEG/MEG decoders.}
	\label{tab:EEG}
	\vspace{8pt}
	\centering
	\renewcommand\arraystretch{1.2}
	\scalebox{0.9}{
		\begin{tabular}{c c c c c}
			\hline
			\textbf{Layer}                    & \textbf{Input Dimensions}     & \textbf{Output Dimensions}   & \textbf{Kernel Size}  & \textbf{Stride}     \\ \cline{1-1}
			\textbf{Temporal Conv}             & $(B, 1, H, W)$                & $(B, k, H_{1}, W)$         & $(1, k_{1})$               & (1, 1)              \\ 
			\textbf{AvgPool 2d}                 & $(B, k, H_{1}, W)$           & $(B, k, H_{2}, W_{2})$     & $(1, k_{2})$               & $(1, s)$              \\ 
			\textbf{Spatial Conv}              & $(B, k, H_{2}, W_{2})$       & $(B, k, H_{2}, W_{2})$     & $(k_{3}, 1)$              & (1, 1)              \\ \cline{1-1}
			\textbf{Flatten\&Projection}                & $(B, k, H_{2}, W_{2})$       & $(B, \text{emb}, H_{2} \cdot W_{2})$ & (1, 1) & (1, 1) \\ \hline
	\end{tabular}}
\end{table}

\subsubsection{Feature selection}

During the feature decoding process, we observed that the high dimensionality of the structural feature $Z_{CLIP}^i$ led to certain dimensions exhibiting low decoding accuracy, which could potentially hinder subsequent structural constraints. To address this issue, we devised a feature selection algorithm (as shown in Algorithm 1) that retains only the top $k\%$ ($k=25$) dimensions of each CLIP feature layer while masking the remaining dimensions.

\begin{figure}[htbp!]
	\centering
	\includegraphics[width=0.95\linewidth]{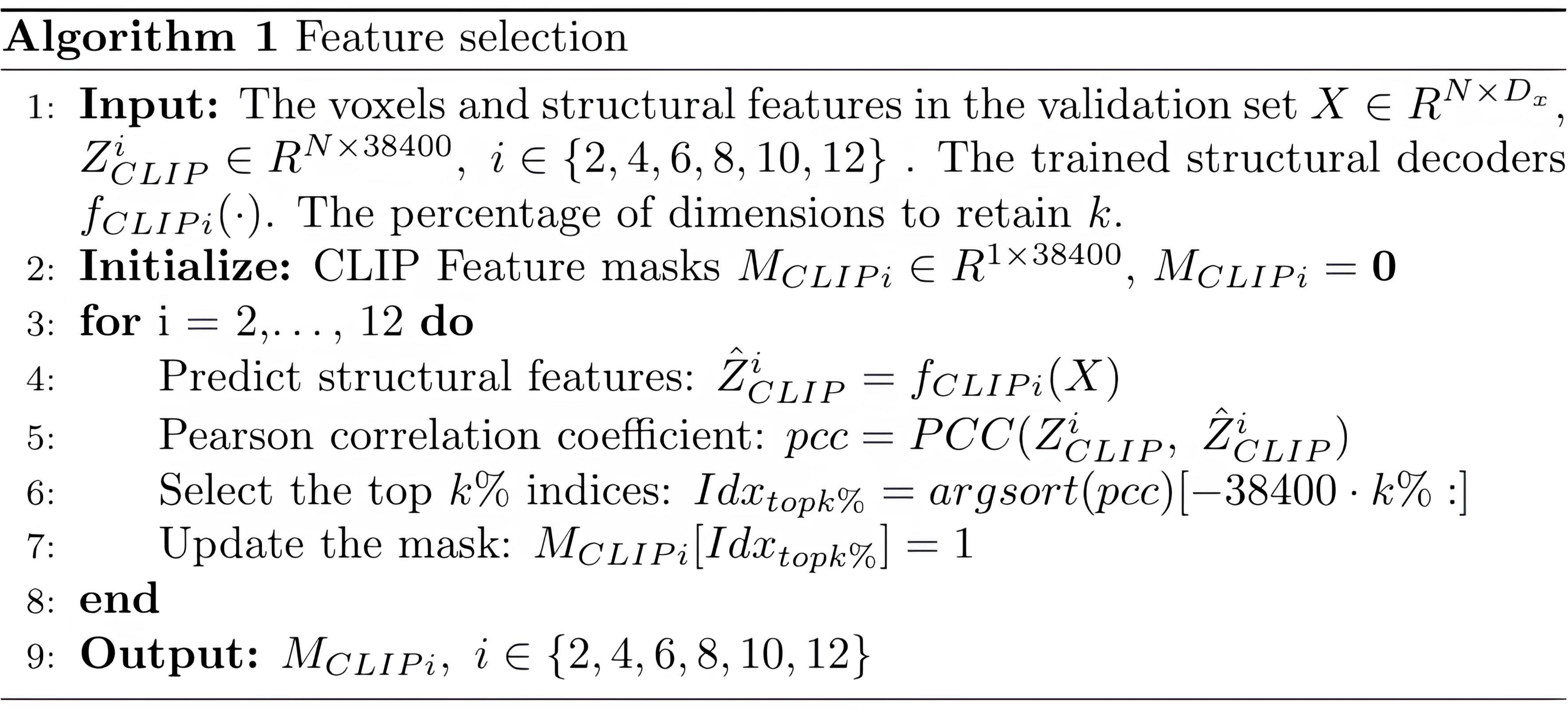}
	\label{fig:algorithmflowchart1}
\end{figure}

\begin{figure}[htbp!]
	\centering
	\includegraphics[width=0.9\linewidth]{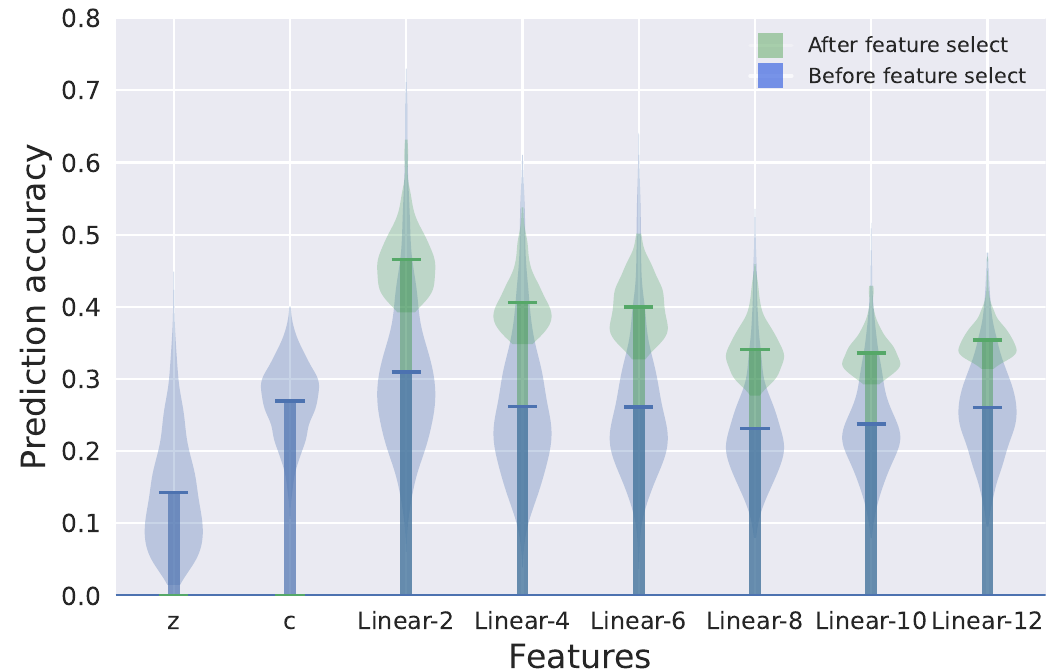}
	\caption{Decoding accuracy of each feature on fMRI data. The results have
		been averaged over 4 subjects. The violin plot illustrates the distribution of decoding accuracy for each feature dimension. The blue bars represent the
		average prediction accuracy of all units in each feature while the green bars represent the average prediction accuracy of structural features after feature selection.}
	\label{fig:decodingaccviolin}
\end{figure}

As illustrated in Fig. \ref{fig:decodingaccviolin}, following the feature selection process, the decoding accuracy of the retained structural features remains consistently at a higher level.

\subsection{Inference}

After decoding the aforementioned features from brain signal, we employed the two-stage inference framework, as illustrated in Algorithm 2, to reconstruct the semantic representation of the stimulus image and align its structural information.
\begin{figure}[htbp!]
	\centering
	\includegraphics[width=1\linewidth]{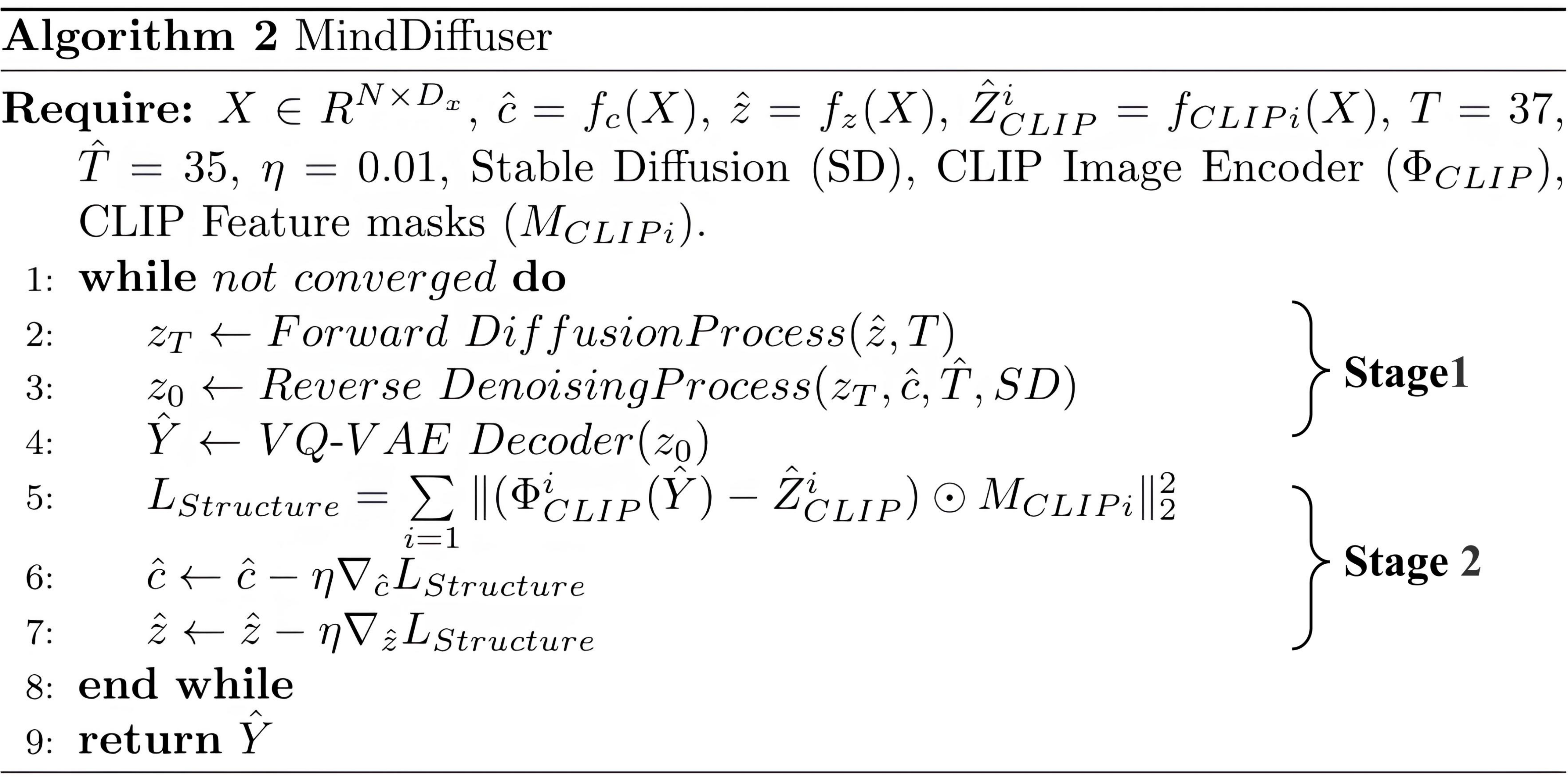}
	\label{fig:algorithmflowchart2}
\end{figure}

\textbf{(1) Stage 1: Semantic information incorporating by reversing the ventral pathway}

The trained $f_c$ and $f_z$ models are employed to decode the CLIP text embeddings $\hat{c} = f_c(X)$ and latent vectors $\hat{z} = f_z(X)$ for the images in the test set. These decoded feature vectors are then input into Stable Diffusion, as shown in Fig. \ref{fig:overview2}(b). To inject image prior information into the latent space, the decoded $\hat{z}$ undergoes a forward diffusion process for T steps, as described in equations \ref{equ1} and \ref{equ2}, which results in the computation of $z_T$.

\begin{equation}\label{equ1}
	q(z_t \mid z_{t-1}) = \mathcal{N}(z_t; \sqrt{\alpha_t}z_{t-1}, (1-\alpha_t)I), \quad t=0,1,\dots, T
\end{equation}

\begin{equation}\label{equ2}
	z_T=\sqrt{\overline{\alpha_T}} z_0+ \sqrt{1-\overline{\alpha_T}}\epsilon        \  and \ z_0=\hat{z}.
\end{equation}

During each reverse denoising iteration, the U-Net \cite{ronneberger2015u} incorporates the decoded CLIP text embedding $\hat{c}$ into $z_T$ using cross-attention, as specified in equation \ref{equ3}.
\begin{equation}
	\label{equ3}
	\begin{split}
		CrossAttention(Q,K,V)=softmax(\frac{Q K^T}{\sqrt{d}}), \\  Q=W_Q^i\cdot\phi_i(z_t), K=W_K^i\cdot \hat{c}, V=W_V^i\cdot \hat{c}.
	\end{split}
\end{equation}
where $\phi_i(z_t)$ represents the middle-layer feature of U-Net, and $W_Q^i$, $W_K^i$, $W_V^i$ denote the pre-trained projection matrixs. The images generated by this process contain semantic information and coarse-grained content.

\textbf{(2) Stage 2: Structural information aligning by reversing the dorsal pathway} 

In Stage 1, the decoded CLIP text embedding $\hat{c}$ and the VQ-VAE latent embedding $\hat{z}$ are utilized to generate an initial reconstructed image $\hat{Y}$ containing coarse-grained semantic information. To further align the structural information of the reconstructed image with the ground truth, we design a structural loss function based on features extracted from the shallow layers of CLIP visual encoder:

\begin{equation}\label{equ5} 
	L_{Structure}=\sum\limits_{i=1}\Vert ( \Phi^i_{CLIP}(\hat{Y})  -\hat{Z}^i_{CLIP}) \odot M_{CLIPi} \Vert_2^2, \end{equation}
where $\Phi^i_{CLIP}(\cdot)$ represents the $i$-th linear layer of the CLIP visual encoder, $M_{CLIPi}$ denotes the $i$-th feature mask derived using Algorithm 1, and $\odot$ indicates the Hadamard product.

As shown in Fig. \ref{fig:overview2}(b),  low-level features are first extracted from the initial reconstructed image using the CLIP visual encoder. The structural loss is then computed as the Mean Squared Error (MSE) between these features and the structural features $\hat{Z}^i_{CLIP}$ decoded from brain signals. Since all components of the model are differentiable, we iteratively optimize $\hat{c}$ and $\hat{z}$ by backpropagating the gradients of this loss function. The refined $\hat{c}$ and $\hat{z}$ are subsequently input into the image generator from the first stage to iteratively update the reconstructed image. This iterative process continues until convergence, ensuring controlled reconstruction of the output.

\section{Experiment results}
\subsection{Dataset Description}

As shown in Table \ref{tab:datasets}, we perform extensive experiments on three large-scale, visually evoked brain response datasets to demonstrate the superiority of our proposed approach.
\begin{table}[htbp!]
	\caption{Details of the datasets used in our experiments.}
	\label{tab:datasets}
	\vspace{8pt}
	\centering
	\renewcommand\arraystretch{1.2}
	\scalebox{1}{
		\begin{tabular}{c|cccc}
			\toprule
			Dataset    & Data type &  Training & Testing & ROIs or Channels \\ \hline
			NSD \cite{allen2022massive}        & fMRI              & 8859     & 982     & 10 ROIs          \\ 
			Things-EEG \cite{gifford2022large} & EEG              & 16540    & 200   & 64 Channels      \\ 
			Things-MEG \cite{hebart2023things} & MEG              & 19848    & 200   & 271 Channels      \\\bottomrule
	\end{tabular}}
\end{table}

\subsubsection{Natural Scenes Dataset (NSD)} The NSD dataset contains high-resolution fMRI data from 8 participants, each viewing 9,000 to 10,000 unique natural scenes with 22,000 to 30,000 repetitions across 30-40 MRI sessions. For our experiments, we used data from participants 1, 2, 5, and 7, who completed all trials. Voxel data from visual cortex ROIs (V1, V2, V3, hV4, VO, PHC, MT, MST, LO, IPS) were extracted for further analysis.

The visual stimuli were sourced from the COCO dataset \cite{lin2014microsoft}, with captions linked via COCO IDs. Each subject's training set included 8,859 stimuli and 24,980 fMRI trials (up to 3 trials per image), while the test set had 982 stimuli and 2,770 trials. For trials with multiple repetitions, we computed the average response to improve the signal-to-noise ratio.

\subsubsection{Things-EEG dataset} The THINGS-EEG dataset \cite{gifford2022large} includes EEG data from 10 subjects performing a visual target detection task using the RSVP paradigm. Each participant completed four experiments, yielding 82,160 trials: 16,540 training trials (repeated 4 times) and 200 testing trials (repeated 80 times). Data were recorded with a 64-channel EEG system at 1000 Hz, then downsampled to 100 Hz and filtered to [0.1, 100] Hz. Baseline correction was applied using the 200 ms pre-stimulus period, and 17 occipital and parietal channels were retained. For the test set, 80 trials per image were averaged to improve the signal-to-noise ratio, while individual trials were kept for the training set.

\subsubsection{Things-MEG dataset}
The THINGS-MEG dataset \cite{hebart2023things} contains 271-channel MEG data from 4 subjects across 12 sessions. The training set consists of 1854 concepts, with 12 images and 1 repetition per concept, while the test set includes 200 concepts with 12 repetitions per image.  MEG data were segmented from 0 to 1000 ms post-stimulus, bandpass filtered ([0.1, 40] Hz), baseline corrected, and downsampled to 200 Hz.

\subsection{Experimental Setup}
\subsubsection{Implementation details}
We employed linear regression models for fMRI decoding with $\alpha = 0.15$, implemented via the PyFastL2LiR library\footnote{Available at https://github.com/KamitaniLab/PyFastL2LiR}. To enhance decoding accuracy, we computed the voxel-feature correlation matrix and retained only the most correlated voxels. Specifically, 250, 350, and 4000 voxels were selected for semantic, structural, and VQ-VAE features, respectively.
For EEG and MEG data, we adjusting the output layer to 38,400 dimensions to align with the CLIP features. Other hyperparameters followed the default settings from Li et al. \cite{li2024visual} We used Vit/B-32 as the CLIP backbone to extract the structural features. In Stage 1, we employed the pre-trained Stable Diffusion V1.4 to generate images, using a 37-step forward diffusion and a 35-step reverse denoising process. In Stage 2, structural information was iteratively optimized using the Adam optimizer with a learning rate of 0.01 for 60 steps. The entire training and inference process was conducted on a single A100 (80GB) GPU, with the random seed fixed at 42.
\subsubsection{Evaluation protocol} 
For the NSD dataset, we followed the default training and test set split. Additionally, we used the last 854 samples from the 8854 training samples as the validation set, and the first 8000 samples as the training set.
For the THINGS-EEG and THINGS-MEG dataset, we adopted the same data splitting strategy as Benchetrit et al. \cite{benchetrit2023brain} and Li et al. \cite{li2024visual}, where 1654 image classes were used for training, with 10\% of the training data randomly selected as the validation set. The remaining 200 image classes were used as the test set, ensuring that the test task was zero-shot.

For quantitative comparison with other methods, we follow the approach in Brain-Diffuser \cite{ozcelik2023natural} and use eight image quality metrics. Low-level properties are evaluated using PixCorr, SSIM \cite{wang2004image}, AlexNet(2), and AlexNet(5) \cite{krizhevsky2012imagenet}, while high-level properties are assessed with Inception \cite{szegedy2016rethinking}, CLIP \cite{radford2021learning}, EffNet-B \cite{tan2019efficientnet}, and SwAV \cite{caron2020unsupervised}.

\subsubsection{Comparison methods}
For model comparison, we selected eleven representative works trained on the NSD dataset, two of which have been integrated into MindEye, as shown in Table \ref{tab:result1}.

To verify whether our proposed framework can be integrated into other methods to enhance their performance, we integrated and tested two GAN-based models (Mind-Reader\footnote{Code are available at https://github.com/sklin93/mind-reader} \cite{lin2022mind} and Gu et al.\footnote{Code are available at https://github.com/zijin-gu/meshconv-decoding} \cite{gu2022decoding}) and three Diffusion-based models (Takagi et al.\footnote{Code are available at https://github.com/yu-takagi/StableDiffusionReconstruction} \cite{takagi2023high}, Brain-Diffuser\footnote{Code are available at https://github.com/ozcelikfu/brain-diffuser} \cite{ozcelik2023natural}, and MindEye\footnote{Code are available at https://medarc-ai.github.io/mindeye/} \cite{scotti2024reconstructing}) on the fMRI dataset (NSD). Additionally, we tested three latest Diffusion-based models (NICE\footnote{Code are available at https://github.com/eeyhsong/NICE-EEG} \cite{song2023decoding}, EEGNetV4\footnote{Code are available at https://huggingface.co/PierreGtch/EEGNetv4} \cite{lawhern2018eegnet}, and ATM\footnote{Code are available at https://github.com/dongyangli-del/EEG\_Image\_decode} \cite{li2024visual}) on the EEG dataset (Things-EEG). All models are open-source, and we reproduced them using the official code, integrating them into our framework for testing, as shown in Table \ref{tab:result2}.

Notably, the original ATM paper used the IP-adapter (SDXL) \cite{ye2023ip} as the generative model, but due to memory constraints, we replaced it with Stable Diffusion V1.4. When reproducing three models on the EEG dataset, the high dimensionality of the semantic features (1x15x768) made direct fitting with EEG data challenging. Therefore, we first conducted classification to obtain category labels and then used the prompt `a picture of [ ]' as the semantic label. The following hyperparameters were used during the iterations: Mind-Reader (lr=0.001, steps=4), Gu et al. (lr=0.0005, steps=60), Brain-Diffuser (lr=0.0001, steps=30), MindEye (lr=0.0001, steps=6), and Takagi et al., NICE, EEGNetV4, ATM (lr=0.01, steps=30).

\subsection{Performance comparison}
\definecolor{myred}{HTML}{FFD2D2}      
\definecolor{myorange}{HTML}{FFE5D2}   
\definecolor{myyellow}{HTML}{FFF5D2}   
\begin{table*}[htbp!]
	\caption{Quantitative comparison of MindDiffuser's reconstruction performance against other models. Ozcelik et al. \cite{ozcelik2022reconstruction}, Mind-Reader \cite{lin2022mind}, Takagi et al. \cite{takagi2023high}, and Gu et al. \cite{gu2022decoding} were not evaluated on all eight metrics; we reproduced their results using their open-source code, while the results for the other methods were taken from their papers. PixCorr represents pixel-wise correlation, SSIM is the Structural Similarity Index Metric, and EffNet-B and SWAV measure feature similarity. The remaining four metrics are based on two-way identification (chance = 50\%). All metrics were averaged across subjects 1, 2, 5, and 7. The best, second-best, and third-best results are highlighted in red, orange, and yellow, respectively.} 
	\label{tab:result1}
	
	\vspace{4pt}
	\centering
	\renewcommand\arraystretch{1.2}
	\scalebox{0.53}{
		\begin{tabular}{ccccc|cccc}
			\toprule
			\multirow{2}{*}{Methods}     & \multicolumn{4}{c}{Low-level}                                                             & \multicolumn{4}{c}{High-level}                                                            \\ 
			& PixCorr$ \uparrow $             & SSIM$ \uparrow $                & AlexNet(2)$ \uparrow $          & AlexNet(5)$ \uparrow $          & InceptionV3$ \uparrow $         & CLIP$ \uparrow $                & EffNet-B$ \downarrow $            & SWAV$ \downarrow $                \\ \hline
			Ozcelik et al. \cite{ozcelik2022reconstruction} {[}IJCNN 2022{]} & 0.126                & 0.135                & 0.689                & 0.812                & 0.764                & 0.763                & 0.879               & 0.579                \\
			Mind-Reader \cite{lin2022mind} {[}NeurIPS 2022{]} & 0.104                & 0.294                & 0.709                & 0.839                & 0.782                & 0.781                & 0.853                & 0.463                \\
			Brain-Diffuser \cite{ozcelik2023natural} {[}Scientific Reports 2023{]}             & 0.254                & \cellcolor{myorange}0.356 & 0.942 & 0.962                & 0.872                & 0.915                & 0.775                & 0.423                \\
			Takagi et al. \cite{takagi2023high} {[}CVPR 2023{]}  & 0.222                   & 0.318                   & 0.830                & 0.830                & 0.760                & 0.770                & 0.916                   & 0.578                   \\
			Gu et al. \cite{gu2022decoding} {[}MIDL 2023{]}         & 0.082                & 0.297                & 0.689                & 0.799                & 0.752                & 0.704                & 0.901                & 0.501                \\
			DREAM \cite{xia2024dream} {[}WACV 2024{]}    & 0.274                & 0.328               & 0.939                & 0.967                & 0.934                &\cellcolor{myyellow} 0.941                & \cellcolor{myorange}0.645                & 0.418 \\
			MindEye \cite{scotti2024reconstructing} {[}NeurIPS 2023{]}     & \cellcolor{myred}0.310 & 0.331                & \cellcolor{myyellow}0.947 & \cellcolor{myyellow}0.978 & 0.938 &0.938 & \cellcolor{myyellow}0.647 & \cellcolor{myorange}0.366  
			\\ 
			BrainCLIP \cite{10858771} {[}TMI 2025{]}     & -       & -                & -       & - & 0.938 &0.830 & 0.907      & -  
			\\
			Psychometry \cite{quan2024psychometry} {[}CVPR 2024{]}     & \cellcolor{myorange}0.297 & 0.304                & \cellcolor{myred}0.964 & \cellcolor{myred}0.986 & \cellcolor{myred}0.958 &\cellcolor{myred}0.968 & \cellcolor{myred}0.628 & \cellcolor{myred}0.345  
			\\
			UMBRAE \cite{xia2024umbrae} {[}ECCV 2024{]}     & \cellcolor{myyellow}0.283       & 0.341                & \cellcolor{myorange}0.955       & 0.970 & 0.917 &0.935 & 0.700      & 0.393  
			\\\hline
			Second Sight \cite{kneeland2023second} {[}CCN 2023{]}        & 0.156                & 0.285                & 0.884                & 0.935                & 0.820                & 0.870                & 0.792                & 0.435                \\
			Ours        & 0.256                & \cellcolor{myyellow}0.344                & 0.852                & 0.843                & 0.784                & 0.791                & 0.884                & 0.551                \\ \hdashline
			BrainRAM+MindEye \cite{xie2024brainram} {[}ACM MM 2023{]}       & 0.176                & 0.342                & 0.899                & 0.957                & 0.926                & \cellcolor{myyellow}0.941                & 0.666                & 0.381                \\
			Second Sight+MindEye \cite{kneeland2023second} {[}CCN 2023{]}        & 0.259                & 0.329                & 0.939                & 0.977                & \cellcolor{myyellow}0.939                & 0.939                & \cellcolor{myorange}0.645                & \cellcolor{myyellow}0.367                \\
			Ours+MindEye                 & 0.278 & \cellcolor{myred}0.370 & \cellcolor{myorange}0.955 & \cellcolor{myorange}0.983 & \cellcolor{myorange}0.945 & \cellcolor{myorange}0.951 &0.706 & 0.468                \\ \bottomrule
	\end{tabular}}
\end{table*}

\begin{figure*}[htbp!]
	\centering
	\includegraphics[width=1.0\linewidth]{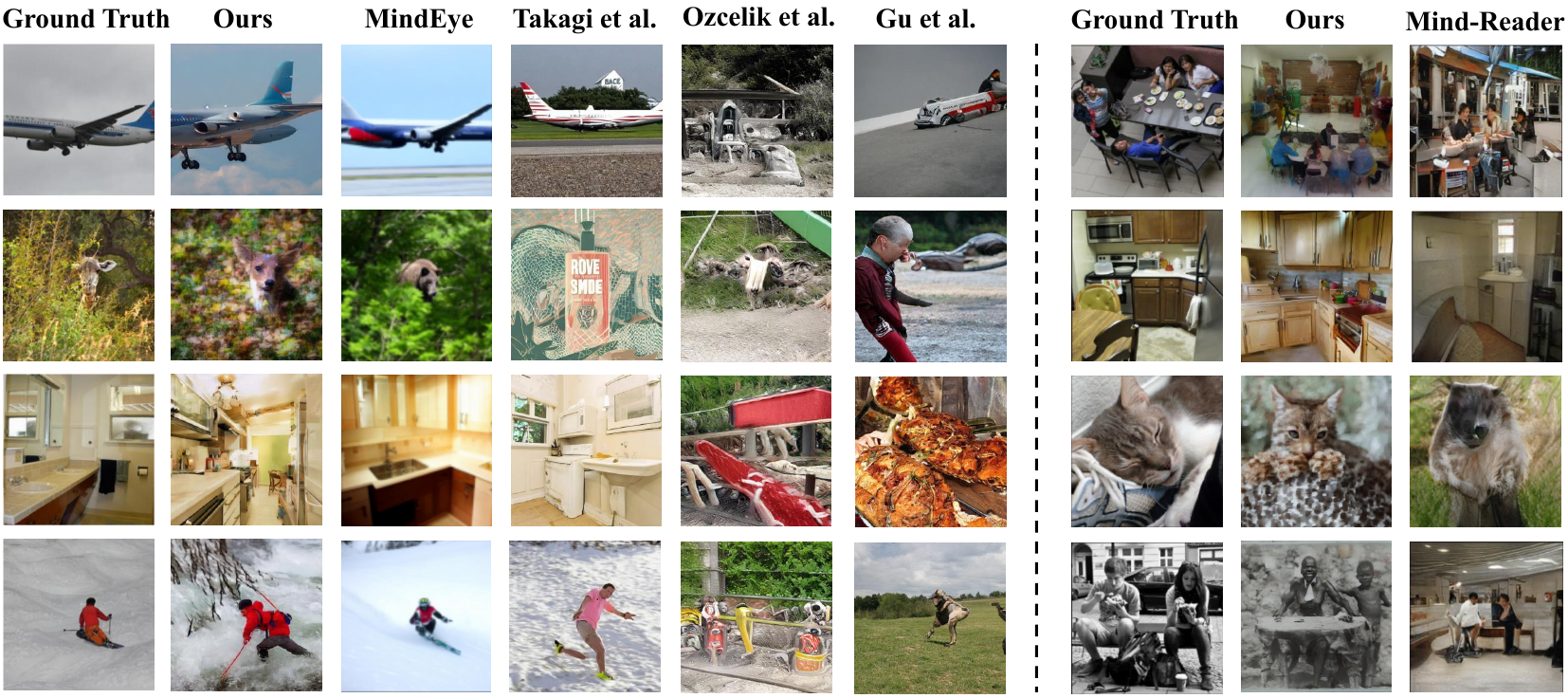}
	\caption{A comparative analysis of reconstruction models on subject 1 of the NSD dataset. The reconstruction results for MindEye and Mind-Reader were taken from their respective papers, while the results for the other methods were reproduced based on the provided code.}
	\label{fig:reconsresults}
\end{figure*}

\begin{figure*}[htbp!]
	\centering
	\setlength{\fboxrule}{0.5pt} 
	\setlength{\fboxsep}{3pt} 
	

	\includegraphics[width=1.0\linewidth]{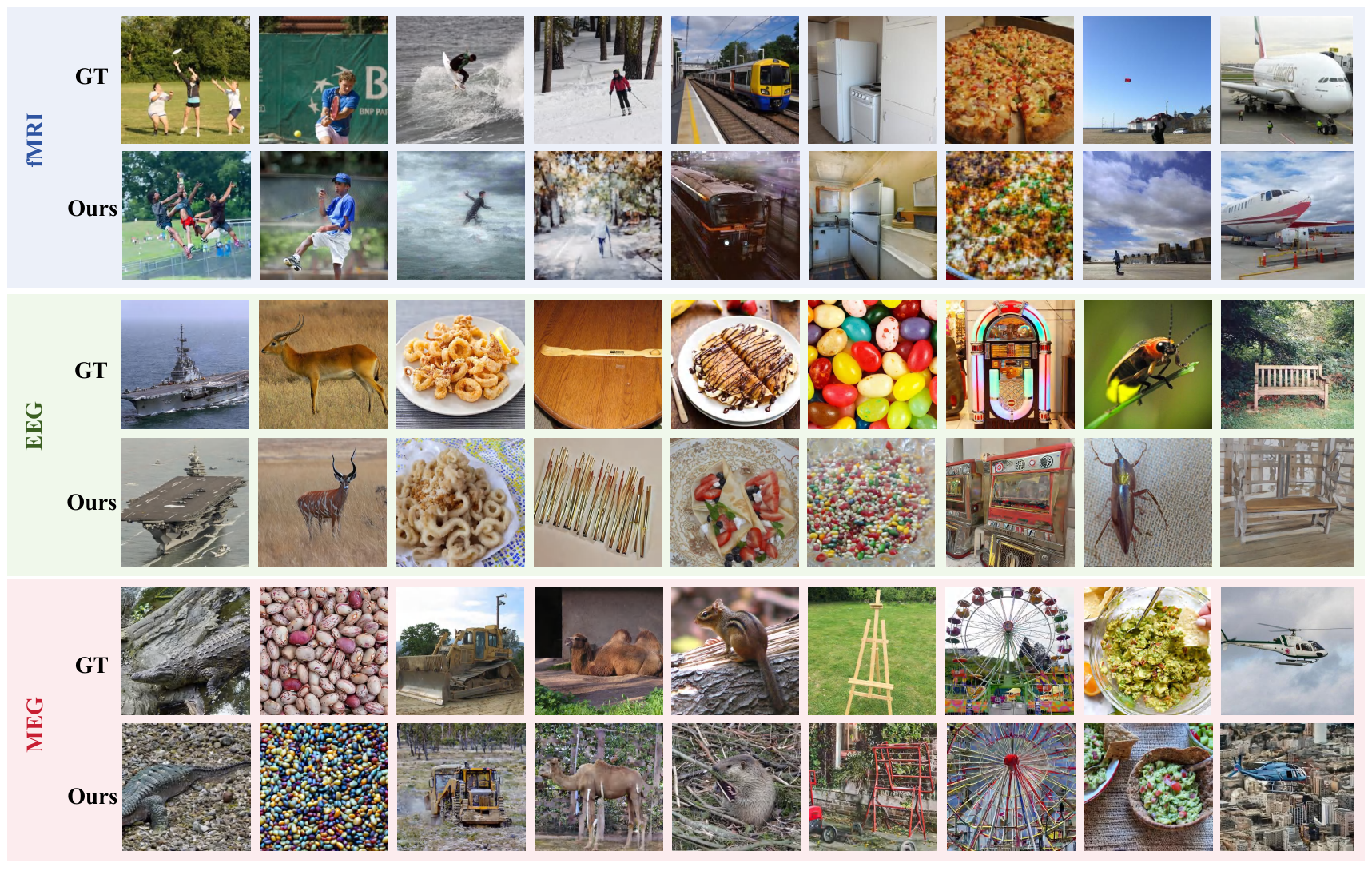}

\caption{Additional reconstruction results from our framework across fMRI, EEG, and MEG modalities.}
	\label{fig:more_recons}
\end{figure*}
Fig. \ref{fig:reconsresults} and Table \ref{tab:result1} present our qualitative and quantitative results, respectively. Fig. \ref{fig:reconsresults} shows the results for subject 1 only, while Table \ref{tab:result1} reports the averages across four subjects. BrainRAM \cite{xie2024brainram} and Second Sight \cite{kneeland2023second} are the multi-stage methods discussed in Section \ref{section_multi_stage}, while the remaining models are single-stage.

As shown in Fig. \ref{fig:reconsresults}, the images reconstructed by our framework exhibit better alignment with the ground truth in terms of structural information, such as spatial position, orientation, and color. Table \ref{tab:result1} shows that, although our method (Ours) is constrained by the performance of the generative model (Stable Diffusion V1.4) and linear decoders, when integrated with the state-of-the-art (SOTA) image reconstruction model MindEye \cite{scotti2024reconstructing} (using Versatile Diffusion \cite{xu2023versatile}) as (Ours+MindEye), it achieves 1 SOTA results and 4 second-best across 8 evaluation metrics. Moreover, our framework outperforms existing multi-stage methods integrated with MindEye in several metrics, with a particularly notable advantage in low-level structural evaluation, demonstrating the effectiveness of the proposed structural alignment method. Additional results are presented in Fig. \ref{fig:more_recons}.

\begin{table*}[htbp!]
	\caption{Versatility of our framework on other methods. We integrated several representative models into the MindDiffuser framework and evaluated its performance on the NSD, Things-EEG and Things-MEG datasets. Results were averaged across different subjects. Since MindEye released reconstruction images for four subjects, we used these images as the starting point for iteration to ensure a fair comparison. Green backgrounds indicate performance improvements, and red backgrounds indicate performance decreases. }
	\label{tab:result2}
	\vspace{4pt}
	\centering
	\renewcommand\arraystretch{1.5}
	\scalebox{0.45}{
		\begin{tabular}{cccccccccccc}
			\toprule
			& \multicolumn{3}{c}{}                                                                                                        & \multicolumn{4}{c}{Low-level}                                                                                                           & \multicolumn{4}{c}{High-level}                                                                                                        \\ \cline{5-12} 
			\multirow{-2}{*}{Data type} & \multicolumn{3}{c}{\multirow{-2}{*}{Methods}}                                                                               & PixCorr$ \uparrow $             & SSIM$ \uparrow $                & AlexNet(2)$ \uparrow $          & AlexNet(5)$ \uparrow $          & InceptionV3$ \uparrow $         & CLIP$ \uparrow $                & EffNet-B$ \downarrow $            & SWAV$ \downarrow $                               \\ \hline
			& \multicolumn{1}{c}{}                                   & \multicolumn{1}{c}{}                                & Base model  & 0.104                            & 0.294                            & 0.709                            & 0.839                           & 0.782                           & 0.781                           & 0.853                           & 0.463                           \\
			& \multicolumn{1}{c}{}                                   & \multicolumn{1}{c}{}                                & + Ours & 0.123                            & 0.320                             & 0.715                            & 0.843                           & 0.836                           & 0.784                           & 0.851                           & 0.459                           \\ \cline{4-12} 
			& \multicolumn{1}{c}{}                                   & \multicolumn{1}{c}{\multirow{-3}{*}{Mind-Reader \cite{lin2022mind}}}   & $\Delta$           & \cellcolor[HTML]{C9E4B4}17.746\% & \cellcolor[HTML]{C9E4B4}8.581\%  & \cellcolor[HTML]{C9E4B4}0.811\%  & \cellcolor[HTML]{C9E4B4}0.417\% & \cellcolor[HTML]{C9E4B4}1.027\% & \cellcolor[HTML]{C9E4B4}0.352\% & \cellcolor[HTML]{C9E4B4}0.059\% & \cellcolor[HTML]{C9E4B4}0.863\% \\ \cline{3-12} 
			& \multicolumn{1}{c}{}                                   & \multicolumn{1}{c}{}                                & Base model  & 0.082                            & 0.297                            & 0.689                            & 0.799                           & 0.752                           & 0.704                           & 0.901                           & 0.501                           \\
			& \multicolumn{1}{c}{}                                   & \multicolumn{1}{c}{}                                & + Ours & 0.120                            & 0.339                            & 0.785                            & 0.845                           & 0.755                           & 0.707                           & 0.892                           & 0.516                           \\ \cline{4-12} 
			& \multicolumn{1}{c}{\multirow{-6}{*}{GAN-based}}        & \multicolumn{1}{c}{\multirow{-3}{*}{Gu et al. \cite{gu2022decoding}}}        & $\Delta$           & \cellcolor[HTML]{C9E4B4}44.848\% & \cellcolor[HTML]{C9E4B4}13.770\% & \cellcolor[HTML]{C9E4B4}13.773\% & \cellcolor[HTML]{C9E4B4}5.695\% & \cellcolor[HTML]{C9E4B4}0.399\% & \cellcolor[HTML]{C9E4B4}0.319\% & \cellcolor[HTML]{C9E4B4}1.026\% & \cellcolor[HTML]{F4B7BE}2.894\% \\ \cline{2-12} 
			& \multicolumn{1}{c}{}                                   & \multicolumn{1}{c}{}                                & Base model  & 0.222                            & 0.318                            & 0.718                            & 0.759                           & 0.700                           & 0.677                           & 0.916                           & 0.578                           \\
			& \multicolumn{1}{c}{}                                   & \multicolumn{1}{c}{}                                & + Ours & 0.249                            & 0.341                            & 0.750                            & 0.770                           & 0.698                           & 0.682                           & 0.884                           & 0.551                           \\ \cline{4-12} 
			& \multicolumn{1}{c}{}                                   & \multicolumn{1}{c}{\multirow{-3}{*}{Takagi et al. \cite{takagi2023high}}}        & $\Delta$           & \cellcolor[HTML]{C9E4B4}12.528\% & \cellcolor[HTML]{C9E4B4}7.154\%  & \cellcolor[HTML]{C9E4B4}4.313\%  & \cellcolor[HTML]{C9E4B4}1.416\% & \cellcolor[HTML]{F4B7BE}0.286\% & \cellcolor[HTML]{C9E4B4}0.701\% & \cellcolor[HTML]{C9E4B4}3.465\% & \cellcolor[HTML]{C9E4B4}4.669\% \\ \cline{3-12} 
			& \multicolumn{1}{c}{}                                   & \multicolumn{1}{c}{}                                & Base model  & 0.194                            & 0.327                            & 0.851                            & 0.882                           & 0.823                           & 0.818                           & 0.854                           & 0.522                           \\
			& \multicolumn{1}{c}{}                                   & \multicolumn{1}{c}{}                                & + Ours & 0.213                            & 0.341                            & 0.863                            & 0.897                           & 0.831                           & 0.831                           & 0.853                           & 0.517                           \\ \cline{4-12} 
			& \multicolumn{1}{c}{}                                   & \multicolumn{1}{c}{\multirow{-3}{*}{Brain-Diffuser \cite{ozcelik2023natural}}} & $\Delta$           & \cellcolor[HTML]{C9E4B4}9.794\%  & \cellcolor[HTML]{C9E4B4}4.358\%  & \cellcolor[HTML]{C9E4B4}1.469\%  & \cellcolor[HTML]{C9E4B4}1.700\% & \cellcolor[HTML]{C9E4B4}0.972\% & \cellcolor[HTML]{C9E4B4}1.620\% & \cellcolor[HTML]{C9E4B4}0.146\% & \cellcolor[HTML]{C9E4B4}0.957\% \\ \cline{3-12} 
			& \multicolumn{1}{c}{}                                   & \multicolumn{1}{c}{}                                & Base model  & 0.259                            & 0.355                            & 0.939                            & 0.965                           & 0.941                           & 0.935                           & 0.723                           & 0.453                           \\
			& \multicolumn{1}{c}{}                                   & \multicolumn{1}{c}{}                                & + Ours & 0.278                            & 0.370                            & 0.955                            & 0.983                           & 0.945                           & 0.951                           & 0.706                           & 0.468                           \\ \cline{4-12} 
			\multirow{-15}{*}{fMRI}     & \multicolumn{1}{c}{}                                   & \multicolumn{1}{c}{\multirow{-3}{*}{MindEye \cite{scotti2024reconstructing}}}       & $\Delta$           & \cellcolor[HTML]{C9E4B4}7.336\%  & \cellcolor[HTML]{C9E4B4}4.231\%  & \cellcolor[HTML]{C9E4B4}1.701\%  & \cellcolor[HTML]{C9E4B4}1.865\% & \cellcolor[HTML]{C9E4B4}0.452\% & \cellcolor[HTML]{C9E4B4}1.711\% & \cellcolor[HTML]{C9E4B4}2.351\% & \cellcolor[HTML]{F4B7BE}3.312\% \\ \cline{1-1} \cline{3-12} 
			& \multicolumn{1}{c}{}                                   & \multicolumn{1}{c}{}                                & Base model  & 0.108                                     & 0.284                            & 0.599                            & 0.631                           & 0.647                           & 0.644                           & 0.949                           & 0.701                                 \\
			& \multicolumn{1}{c}{}                                   & \multicolumn{1}{c}{}                                & + Ours & 0.114                                     & 0.293                            & 0.610                            & 0.641                           & 0.651                           & 0.647                           & 0.924                           & 0.688                             \\ \cline{4-12} 
			& \multicolumn{1}{c}{}                                   & \multicolumn{1}{c}{\multirow{-3}{*}{NICE \cite{song2023decoding}}}          & $\Delta$           & \cellcolor[HTML]{C9E4B4}5.554\%           & \cellcolor[HTML]{C9E4B4}3.169\%  & \cellcolor[HTML]{C9E4B4}1.836\%  & \cellcolor[HTML]{C9E4B4}1.643\% & \cellcolor[HTML]{C9E4B4}0.621\% & \cellcolor[HTML]{C9E4B4}0.528\% & \cellcolor[HTML]{C9E4B4}2.634\% & \cellcolor[HTML]{C9E4B4}1.854\% \\ \cline{3-12} 
			& \multicolumn{1}{c}{}                                   & \multicolumn{1}{c}{}                                & Base model  & 0.124                                     & 0.313                            & 0.582                            & 0.624                           & 0.638                           & 0.653                           & 0.942                           & 0.693                           \\
			& \multicolumn{1}{c}{}                                   & \multicolumn{1}{c}{}                                & + Ours & 0.131                                     & 0.324                            & 0.593                            & 0.629                           & 0.631                           & 0.655                           & 0.931                           & 0.680                            \\ \cline{4-12} 
			& \multicolumn{1}{c}{}                                   & \multicolumn{1}{c}{\multirow{-3}{*}{EEGNetV4 \cite{lawhern2018eegnet}}}          & $\Delta$           & \cellcolor[HTML]{C9E4B4}5.645\%          & \cellcolor[HTML]{C9E4B4}3.514\%  & \cellcolor[HTML]{C9E4B4}1.931\%  & \cellcolor[HTML]{C9E4B4}0.887\% & \cellcolor[HTML]{F4B7BE}1.097\% & \cellcolor[HTML]{C9E4B4}0.313\% & \cellcolor[HTML]{C9E4B4}1.168\% & \cellcolor[HTML]{C9E4B4}1.876\% \\ \cline{3-12} 
			& \multicolumn{1}{c}{}                                   & \multicolumn{1}{c}{}                                & Base model  & 0.117                                     & 0.397                            & 0.597                            & 0.655                           & 0.630                           & 0.674                           & 0.932                           & 0.687                           \\
			& \multicolumn{1}{c}{}                                   & \multicolumn{1}{c}{}                                & + Ours & 0.128                                     & 0.416                            & 0.609                            & 0.666                           & 0.626                           & 0.677                           & 0.926                           & 0.675                           \\ \cline{4-12} 
			\multirow{-9}{*}{EEG}       & \multicolumn{1}{c}{}                                   & \multicolumn{1}{c}{\multirow{-3}{*}{ATM \cite{li2024visual}}}           & $\Delta$           & \cellcolor[HTML]{C9E4B4}9.402\%          & \cellcolor[HTML]{C9E4B4}4.685\%  & \cellcolor[HTML]{C9E4B4}2.076\%  & \cellcolor[HTML]{C9E4B4}1.679\% & \cellcolor[HTML]{F4B7BE}0.730\% & \cellcolor[HTML]{C9E4B4}0.416\% & \cellcolor[HTML]{C9E4B4}0.623\% & \cellcolor[HTML]{C9E4B4}1.862\% \\ \cline{1-1} \cline{3-12} 
			& \multicolumn{1}{c}{}                                   & \multicolumn{1}{c}{}                                & Base model  & 0.010                            & 0.222                            & 0.540                            & 0.615                           & 0.627                           & 0.653                           & 0.949                           & 0.706                           \\
			& \multicolumn{1}{c}{}                                   & \multicolumn{1}{c}{}                                & + Ours & 0.016                            & 0.233                            & 0.580                            & 0.658                           & 0.639                           & 0.663                           & 0.945                           & 0.696                           \\ \cline{4-12} 
			& \multicolumn{1}{c}{}                                   & \multicolumn{1}{c}{\multirow{-3}{*}{ATM \cite{li2024visual}}}           & $\Delta$           & \cellcolor[HTML]{C9E4B4}60.000\%                         & \cellcolor[HTML]{C9E4B4}4.955\%                          & \cellcolor[HTML]{C9E4B4}7.407\%                          & \cellcolor[HTML]{C9E4B4}6.992\%                         & \cellcolor[HTML]{C9E4B4}1.914\%                         & \cellcolor[HTML]{C9E4B4}1.531\%                         & \cellcolor[HTML]{C9E4B4}0.421\%                        & \cellcolor[HTML]{C9E4B4}1.416\%                        \\ \cline{3-12} 
			& \multicolumn{1}{c}{}                                   & \multicolumn{1}{c}{}                                & Base model  & 0.020                            & 0.247                            & 0.530                            & 0.555                           & 0.538                           & 0.557                           & 0.980                           & 0.734                           \\
			& \multicolumn{1}{c}{}                                   & \multicolumn{1}{c}{}                                & + Ours & 0.027                            & 0.260                            & 0.585                            & 0.625                           & 0.560                           & 0.575                           & 0.973                           & 0.724                           \\ \cline{4-12} 
			\multirow{-6}{*}{MEG}       & \multicolumn{1}{c}{\multirow{-24}{*}{Diffusion-based}} & \multicolumn{1}{c}{\multirow{-3}{*}{Benchetrit \cite{benchetrit2023brain}}}          & $\Delta$           & \cellcolor[HTML]{C9E4B4}35.000\%                         & \cellcolor[HTML]{C9E4B4}5.263\%                          & \cellcolor[HTML]{C9E4B4}10.377\%                         & \cellcolor[HTML]{C9E4B4}12.613\%                        & \cellcolor[HTML]{C9E4B4}4.089\%                         & \cellcolor[HTML]{C9E4B4}3.232\%                         & \cellcolor[HTML]{C9E4B4}0.714\%                        & \cellcolor[HTML]{C9E4B4}1.362\%                        \\ \bottomrule
	\end{tabular}}
	
\end{table*}

\subsection{Versatility on other methods}

\begin{figure}[htbp!]
	\centering
	\setlength{\fboxrule}{0.5pt} 
	\setlength{\fboxsep}{3pt} 
	

			\includegraphics[width=1.0\linewidth]{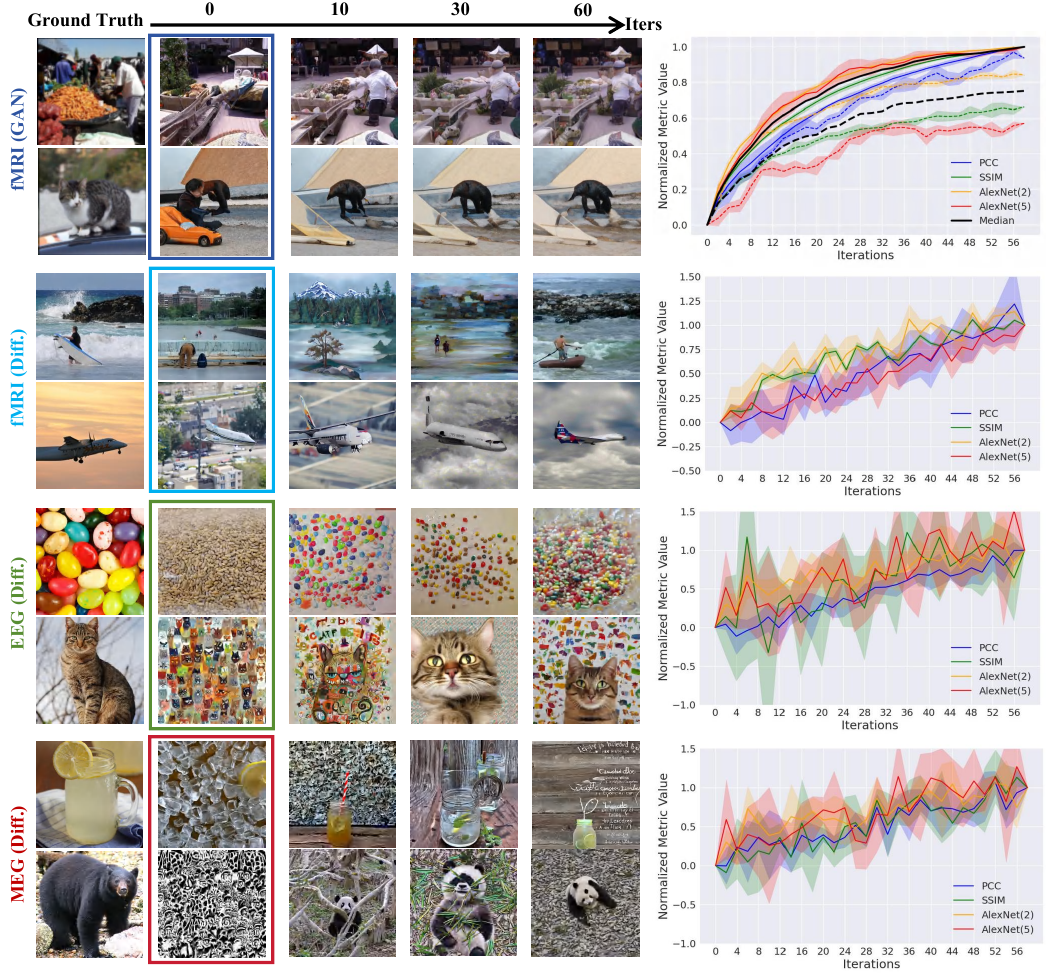}

	
\caption{Qualitative results showing MindDiffuser's performance improvement on different models. Four representative methods are selected (from top to bottom): Gu et al. \cite{gu2022decoding} on fMRI, Takagi et al. \cite{takagi2023high} on fMRI, and ATM \cite{li2024visual} on the EEG and MEG datasets. The initial reconstructions of these methods are highlighted in dark blue, blue, green, and red, respectively. The plots on the right show the progression of low-level metrics for four test subjects, where 0 represents the initial result and 1 denotes the result after 60 iterations. All metrics are normalized to the range.}
	\label{fig:iter4}
\end{figure}


To verify whether integrating our proposed framework into other pre-trained image reconstruction models improves their reconstruction performance during inference, we selected ten representative methods from different data modalities (fMRI, EEG and MEG) and generative model types (GAN, Diffusion). These methods were experimentally evaluated using their official open-source code, and the results averaged across multiple subjects are presented in Table \ref{tab:result2}.

The following observations and conclusions can be drawn from the table:
\begin{itemize}
	\item Our framework effectively enhances the low-level structural metrics of all models, confirming the versatility of our proposed method in structural alignment.
	
	\item Our method shows minimal improvement in high-level semantic metrics, and in some cases, even leads to a decrease. This is because we do not impose semantic-level constraints as done by Kneeland et al. \cite{kneeland2023second} or Xie et al. \cite{xie2024brainram}
	
	\item Our framework achieves a greater improvement in low-level metrics on fMRI and MEG data compared to EEG data. This is likely due to the significantly lower spatial resolution of EEG data, which makes it challenging to decode fine-grained structural features (as evidenced by the results in Fig. \ref{fig:visualcortexsub1}(e)).

\end{itemize}

Fig. \ref{fig:iter4} illustrates the iterative refinement of reconstruction results from four representative methods. A clear trend is observed where, with an increasing number of iterations, the reconstructions progressively align with the ground truth in terms of structural fidelity. For instance, in the sixth row, the image evolves from abstract, cat-like shapes into a well-defined tabby cat. This process also demonstrates that structural constraints can effectively mitigate semantic decoding biases, as evidenced in the fifth row, where an initial reconstruction resembling a pile of beans is corrected to match the ground truth of rainbow-colored candies. However, the method does not guarantee perfect alignment of all details. In the fourth row, while the orientation of the airplane correctly matches the ground truth, its background color still shows a significant deviation. Addressing this limitation will be a key focus of our future work.

Further analysis reveals distinct behaviors across different models and modalities.
\begin{itemize}
\item Comparison of Generative Models: The GAN-based reconstruction exhibits rapid initial convergence, with its low-level metrics quickly reaching a plateau in the early iterations. In contrast, the process for diffusion-based methods is less smooth, characterized by a fluctuating yet consistently upward trend. Notably, their low-level metrics show no signs of saturation, suggesting that further performance gains are achievable with greater computational investment.

\item Comparison of Data Modalities: Although our framework proves effective across all tested modalities, the magnitude of fluctuation in the metrics varies. Specifically, fMRI, MEG, and EEG show progressively larger oscillations. We hypothesize that this phenomenon is attributable to the inherent differences in their spatial resolutions. Modalities with lower spatial resolution, such as MEG and EEG, capture fewer low-level features from the source \cite{buzsaki2012origin, henson2009selecting}, potentially leading to greater instability during the iterative refinement process.
\end{itemize}

\subsection{Ablation study and Parameter sensitivity analysis}
In this section, we present a series of ablation studies and analyses to evaluate the key components of our model. First, we conducted ablation studies on different Regions of Interest (ROIs) at the training stage and on various feature combinations during inference. Next, we performed a sensitivity analysis on our feature selection mechanism, specifically examining the impact of the feature retention ratio, k\%. Finally, to validate our architectural choice for feature fitting, we replaced the linear regression model with non-linear alternatives—a multi-layer perceptron (MLP) and a Transformer—and investigated the resulting effect on performance.

\subsubsection{Ablation study of different ROIs}
The visual cortex ROIs are functionally divided into low-level visual cortex (LVC) and high-level visual cortex (HVC). The LVC, including areas such as V1, V2, V3, and V3ab, primarily processes basic sensory inputs, such as color, edges, and textures, which are texture features. In contrast, the HVC, including areas like VO, PHC, LO, and MT, is involved in processing more abstract and semantic features. 
\begin{figure}[htbp!]
	\centering
	\includegraphics[width=0.8\linewidth]{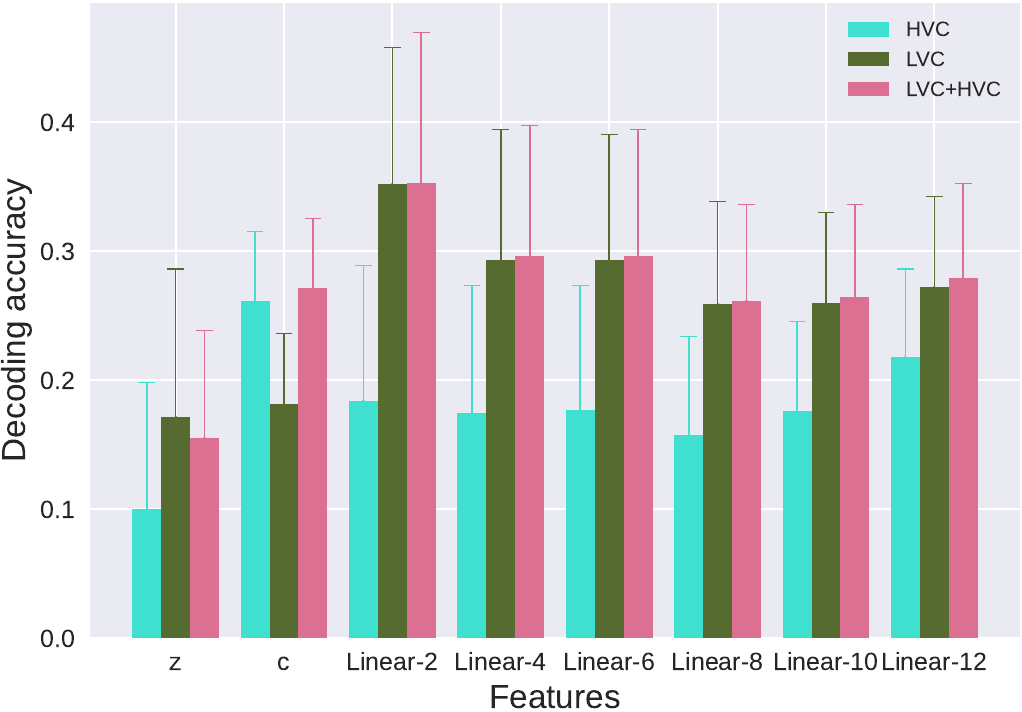}
	\caption{The decoding accuracy of different ROIs during the decoding phase. All experiments were conducted on subject 1. The figure shows the standard deviation for each dimension.}
	\label{fig:ablationforvcbar}
\end{figure}

Initially, we used all ROIs of the visual cortex (LVC + HVC) for feature decoding. To investigate the importance of different ROIs in decoding various features, we separately performed decoding using only LVC and only HVC. The decoding accuracies are shown in Fig. \ref{fig:ablationforvcbar}.  As illustrated in this figure, when decoding the semantic feature $c$, the HVC plays a dominant role, with the removal of LVC having minimal effect on decoding performance. In contrast, for decoding structural features (e.g., Linear-2, Linear-3, $\cdots$, Linear-12), LVC is the primary contributor, and the removal of HVC does not significantly impact the decoding accuracy. These observations suggest a strong correspondence between the extracted semantic and structural features and the respective functional regions of the brain, thereby validating the efficacy of our feature extraction approach. Notably, when decoding the VQ-VAE latent space feature $z$, LVC again prevails, and the inclusion of HVC even leads to a detrimental effect, indicating that $z$ contains minimal semantic information.

\begin{table}[htbp!]
	\caption{Quantitative comparison of ablation experiment results on different ROIs. }
	\label{tab:ablation2}
	\vspace{8pt}
	\centering
	\renewcommand\arraystretch{1.4}
	\setlength{\tabcolsep}{5pt} 
	\scalebox{1}{
		\begin{tabular}{cc|ccc}
			\toprule
			\multicolumn{2}{c|}{Metric}                                     & w/o HVC & w/o LVC & Full model     \\ \hline
			\multicolumn{1}{c|}{\multirow{4}{*}{Low-level}}  & PixCorr$ \uparrow $     & .051 $\pm$ .012   & .030 $\pm$ .007   & \textbf{.256 $\pm$ .025} \\
			\multicolumn{1}{c|}{}                            & SSIM$ \uparrow $        & .338 $\pm$ .017   & .219 $\pm$ .015   & \textbf{.344 $\pm$ .014} \\
			\multicolumn{1}{c|}{}                            & AlexNet(2)$ \uparrow $  & .811 $\pm$ .021   & .776 $\pm$ .027   & \textbf{.852 $\pm$ .026} \\
			\multicolumn{1}{c|}{}                            & AlexNet(5)$ \uparrow $  & .809 $\pm$ .023   & .795 $\pm$ .019   & \textbf{.843 $\pm$ .023} \\ \hline
			\multicolumn{1}{c|}{\multirow{4}{*}{High-level}} & InceptionV3$ \uparrow $ & .680 $\pm$ .024   & .673 $\pm$ .022   & \textbf{.784 $\pm$ .022} \\
			\multicolumn{1}{c|}{}                            & CLIP$ \uparrow $        & .552 $\pm$ .016   & .554 $\pm$ .014   & \textbf{.791 $\pm$ .021} \\
			\multicolumn{1}{c|}{}                            & EffNet-B$ \downarrow $    & .943 $\pm$ .013   & .921 $\pm$ .030   & \textbf{.884 $\pm$ .037} \\
			\multicolumn{1}{c|}{}                            & SWAV$ \downarrow $       & .612 $\pm$ .024   & .591 $\pm$ .028   & \textbf{.551 $\pm$ .031} \\ \bottomrule
	\end{tabular}}
	
\end{table}

Image reconstruction using features from different ROIs is shown in Table \ref{tab:ablation2}. Removing either LVC or HVC decreases all metrics, with a greater impact on low-level metrics when LVC is removed. This indicates that both LVC and HVC contribute to image reconstruction, although their importance varies depending on the type of features being decoded.

\subsubsection{Ablation study on different feature combinations during inference}

\begin{table}[htbp!]
	\caption{Quantitative comparison of ablation experiment results on different features. }
	\label{tab:ablation1}
	\vspace{8pt}
	\centering
	\renewcommand\arraystretch{1.4}
	\setlength{\tabcolsep}{3pt} 
	\scalebox{1}{
		\begin{tabular}{cc|cccc}
			\toprule
			\multicolumn{2}{c|}{Metric}                                     & w/o $c$ & w/o $z$ & w/o $Z_{CLIP}$ & Full model     \\ \hline
			\multicolumn{1}{c|}{\multirow{4}{*}{Low-level}}  & PixCorr$ \uparrow $     & .218 $\pm$ .013 & .066 $\pm$ .008 & .183 $\pm$ .021       & \textbf{.256 $\pm$ .025} \\
			\multicolumn{1}{c|}{}                            & SSIM$ \uparrow $        & .346 $\pm$ .016 & .292 $\pm$ .017 & .253 $\pm$ .021       & \textbf{.344 $\pm$ .014} \\
			\multicolumn{1}{c|}{}                            & AlexNet(2)$ \uparrow $  & .769 $\pm$ .026 & .811 $\pm$ .024 & .826 $\pm$ .021       & \textbf{.852 $\pm$ .026} \\
			\multicolumn{1}{c|}{}                            & AlexNet(5)$ \uparrow $  & .793 $\pm$ .016 & .809 $\pm$ .025 & .821 $\pm$ .013        & \textbf{.843 $\pm$ .023} \\ \hline
			\multicolumn{1}{c|}{\multirow{4}{*}{High-level}} & InceptionV3$ \uparrow $ & .768 $\pm$ .024 & .763 $\pm$ .018 & .772 $\pm$ .013       & \textbf{.784 $\pm$ .022} \\
			\multicolumn{1}{c|}{}                            & CLIP$ \uparrow $        & .549 $\pm$ .023 & .616 $\pm$ .021 & .597 $\pm$ .022       & \textbf{.791 $\pm$ .021} \\
			\multicolumn{1}{c|}{}                            & EffNet-B$ \downarrow $    & .915 $\pm$ .011 & .901 $\pm$ .024 & .896 $\pm$ .012       & \textbf{.884 $\pm$ .037} \\
			\multicolumn{1}{c|}{}                            & SWAV$ \downarrow $        & .617 $\pm$ .032 & .609 $\pm$ .025 & .574 $\pm$ .017       & \textbf{.551 $\pm$ .031} \\ \bottomrule
	\end{tabular}}
	
\end{table}

We investigated the impact of semantic $c$, structural $Z_{CLIP}$, and VQ-VAE latent space features $z$ on image reconstruction, as shown in Table \ref{tab:ablation1}. The results reveal that removing any of these features degrades reconstruction performance. Specifically, removing $c$ primarily affects high-level metrics, while omitting $Z_{CLIP}$ or $z$ predominantly impacts low-level metrics. This highlights the importance of all three feature types in image reconstruction.

\subsubsection{Sensitivity analysis of the parameter $k\%$ for feature selection}

\begin{figure}[htbp!]
	\centering
	\includegraphics[width=0.9\linewidth]{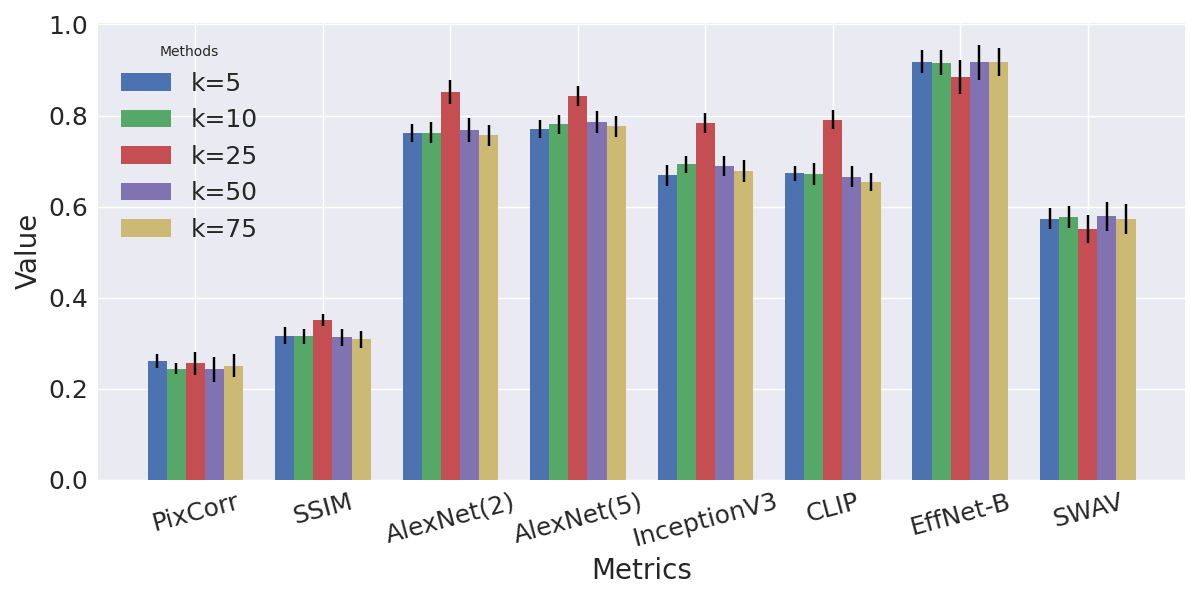}
	\caption{The impact of different structural feature retention ratios ($k\%$) on reconstruction performance. Lower values are better for EffNet-B and SWAV, while higher values are preferred for other metrics. Results are presented as the mean and standard deviation across four subjects.}
	\label{fig:diffk}
\end{figure}

In this section, we analyze the effect of varying proportions of retained structural features on reconstruction performance, as shown in Fig. \ref{fig:diffk}. The results indicate that retaining 25\% of the features yields the best performance across all metrics. Retaining too many features may introduce decoding errors, while retaining too few may fail to preserve sufficient structural information for alignment, both of which degrade reconstruction quality.

\subsubsection{Effect of Model Choice for Feature Fitting}

To evaluate the impact of different feature-fitting architectures on the final reconstruction performance, we replaced the baseline linear regression model with a 3-layer MLP and a 3-layer Transformer. These experiments were conducted on representative models across the fMRI, EEG, and MEG datasets, with the results summarized in Table \ref{tab:result_subset_simplified}.

The results indicate that substituting the linear model with more powerful non-linear architectures, such as the MLP or Transformer, leads to substantial improvements across nearly all metrics. Notably, even metrics that showed a slight decline with the linear model exhibited comprehensive gains after the replacement. This demonstrates that our framework is compatible with various fitting models and that enhanced fitting capacity directly translates to improved image reconstruction performance. However, to preserve model interpretability, we deliberately chose the linear model, accepting a trade-off in performance for clearer neuroscientific insights.

\begin{table}[htbp!]
	\caption{Performance comparison of different model choices for the feature fitting. Results are reported for three representative models, with the best performance for each metric highlighted in bold. Green backgrounds indicate performance improvements, and red backgrounds indicate performance decreases.}
	\label{tab:result_subset_simplified}
	\vspace{8pt}
	\centering
	

		\scalebox{0.55}{ 
			\begin{tabular}{cccccccccccc}
				\toprule
				& \multicolumn{3}{c}{}                                                                                                        & \multicolumn{4}{c}{Low-level}                                                                                                           & \multicolumn{4}{c}{High-level}                                                                                                        \\ \cline{5-12} 
				\multirow{-2}{*}{Data type} & \multicolumn{3}{c}{\multirow{-2}{*}{Methods}}                                                                               & PixCorr$ \uparrow $             & SSIM$ \uparrow $                & AlexNet(2)$ \uparrow $          & AlexNet(5)$ \uparrow $          & InceptionV3$ \uparrow $         & CLIP$ \uparrow $                & EffNet-B$ \downarrow $            & SWAV$ \downarrow $                               \\ \hline
				
				&                                    &                                                          & $\Delta$ (Linear)          & \cellcolor[HTML]{C9E4B4}44.848\% & \cellcolor[HTML]{C9E4B4}\textbf{13.770\%} & \cellcolor[HTML]{C9E4B4}13.773\% & \cellcolor[HTML]{C9E4B4}5.695\% & \cellcolor[HTML]{C9E4B4}0.399\% & \cellcolor[HTML]{C9E4B4}0.319\% & \cellcolor[HTML]{C9E4B4}1.026\% & \cellcolor[HTML]{F4B7BE}2.894\% \\
				&                                    &                                                          & $\Delta$  (MLP)         & \cellcolor[HTML]{C9E4B4}\textbf{46.367\%} & \cellcolor[HTML]{C9E4B4}12.361\% & \cellcolor[HTML]{C9E4B4}\textbf{15.615\%} & \cellcolor[HTML]{C9E4B4}5.889\% & \cellcolor[HTML]{C9E4B4}\textbf{0.536\%} & \cellcolor[HTML]{C9E4B4}0.323\% & \cellcolor[HTML]{C9E4B4}1.643\% & \cellcolor[HTML]{C9E4B4}\textbf{1.763\%} \\
				\multirow{-3}{*}{fMRI}               &                                    & \multirow{-3}{*}{Gu et al. \cite{gu2022decoding}}        & $\Delta$ (Transformer)          & \cellcolor[HTML]{C9E4B4}45.527\% & \cellcolor[HTML]{C9E4B4}13.156\% & \cellcolor[HTML]{C9E4B4}14.236\% & \cellcolor[HTML]{C9E4B4}\textbf{6.472\%} & \cellcolor[HTML]{C9E4B4}0.348\% & \cellcolor[HTML]{C9E4B4}\textbf{0.368\%} & \cellcolor[HTML]{C9E4B4}\textbf{1.733\%} & \cellcolor[HTML]{C9E4B4}1.342\% \\
				\midrule
				
				&                                    &                                                          & $\Delta$ (Linear)            & \cellcolor[HTML]{C9E4B4}9.402\%          & \cellcolor[HTML]{C9E4B4}4.685\%  & \cellcolor[HTML]{C9E4B4}2.076\%  & \cellcolor[HTML]{C9E4B4}1.679\% & \cellcolor[HTML]{F4B7BE}0.730\% & \cellcolor[HTML]{C9E4B4}0.416\% & \cellcolor[HTML]{C9E4B4}0.623\% & \cellcolor[HTML]{C9E4B4}1.862\% \\
				&                                    &                                                          & $\Delta$   (MLP)          & \cellcolor[HTML]{C9E4B4}10.536\%          & \cellcolor[HTML]{C9E4B4}\textbf{5.725\%}  & \cellcolor[HTML]{C9E4B4}3.283\%  & \cellcolor[HTML]{C9E4B4}\textbf{1.784}\% & \cellcolor[HTML]{C9E4B4}1.421\% & \cellcolor[HTML]{C9E4B4}0.675\% & \cellcolor[HTML]{C9E4B4}\textbf{0.856\%} & \cellcolor[HTML]{C9E4B4}2.343\% \\
				\multirow{-3}{*}{EEG}                &                                    & \multirow{-3}{*}{ATM \cite{li2024visual}}               & $\Delta$ (Transformer)           & \cellcolor[HTML]{C9E4B4}\textbf{12.648\%}          & \cellcolor[HTML]{C9E4B4}4.374\%  & \cellcolor[HTML]{C9E4B4}\textbf{3.468\%}  & \cellcolor[HTML]{C9E4B4}1.743\% & \cellcolor[HTML]{C9E4B4}\textbf{1.457\%} & \cellcolor[HTML]{C9E4B4}\textbf{0.723\%} & \cellcolor[HTML]{C9E4B4}0.685\% & \cellcolor[HTML]{C9E4B4}\textbf{2.453\%} \\
				\midrule
				
				&                                    &                                                          & $\Delta$  (Linear)           & \cellcolor[HTML]{C9E4B4}60.000\%                         & \cellcolor[HTML]{C9E4B4}4.955\%                          & \cellcolor[HTML]{C9E4B4}7.407\%                          & \cellcolor[HTML]{C9E4B4}6.992\%                         & \cellcolor[HTML]{C9E4B4}1.914\%                         & \cellcolor[HTML]{C9E4B4}1.531\%                         & \cellcolor[HTML]{C9E4B4}0.421\%                        & \cellcolor[HTML]{C9E4B4}1.416\%                        \\
				&                                    &                                                          & $\Delta$    (MLP)         & \cellcolor[HTML]{C9E4B4}67.534\%                         & \cellcolor[HTML]{C9E4B4}\textbf{6.357\%}                          & \cellcolor[HTML]{C9E4B4}7.684\%                          & \cellcolor[HTML]{C9E4B4}8.312\%                         & \cellcolor[HTML]{C9E4B4}\textbf{2.335\% }                        & \cellcolor[HTML]{C9E4B4}1.625\%                         & \cellcolor[HTML]{C9E4B4}0.342\%                        & \cellcolor[HTML]{C9E4B4}1.523\%                        \\
				\multirow{-3}{*}{MEG}                &                                    & \multirow{-3}{*}{ATM \cite{li2024visual}}               & $\Delta$ (Transformer)           & \cellcolor[HTML]{C9E4B4}\textbf{68.413\%}                         & \cellcolor[HTML]{C9E4B4}5.973\%                          & \cellcolor[HTML]{C9E4B4}\textbf{8.305\%}                          & \cellcolor[HTML]{C9E4B4}\textbf{8.753\%}                         & \cellcolor[HTML]{C9E4B4}1.875\%                         & \cellcolor[HTML]{C9E4B4}\textbf{1.709\% }                        & \cellcolor[HTML]{C9E4B4}\textbf{0.467\% }                       & \cellcolor[HTML]{C9E4B4}\textbf{1.605\% }                       \\ \bottomrule
			\end{tabular}%
		}
\end{table}

\subsection{Model Interpretation}

To investigate how the different features involved in our framework are represented in the brain across spatial and temporal dimensions, we conducted spatial analysis on fMRI data and temporal analysis on EEG and MEG data.

\begin{figure*}[htbp!]
	\centering
	\includegraphics[width=1.0\linewidth]{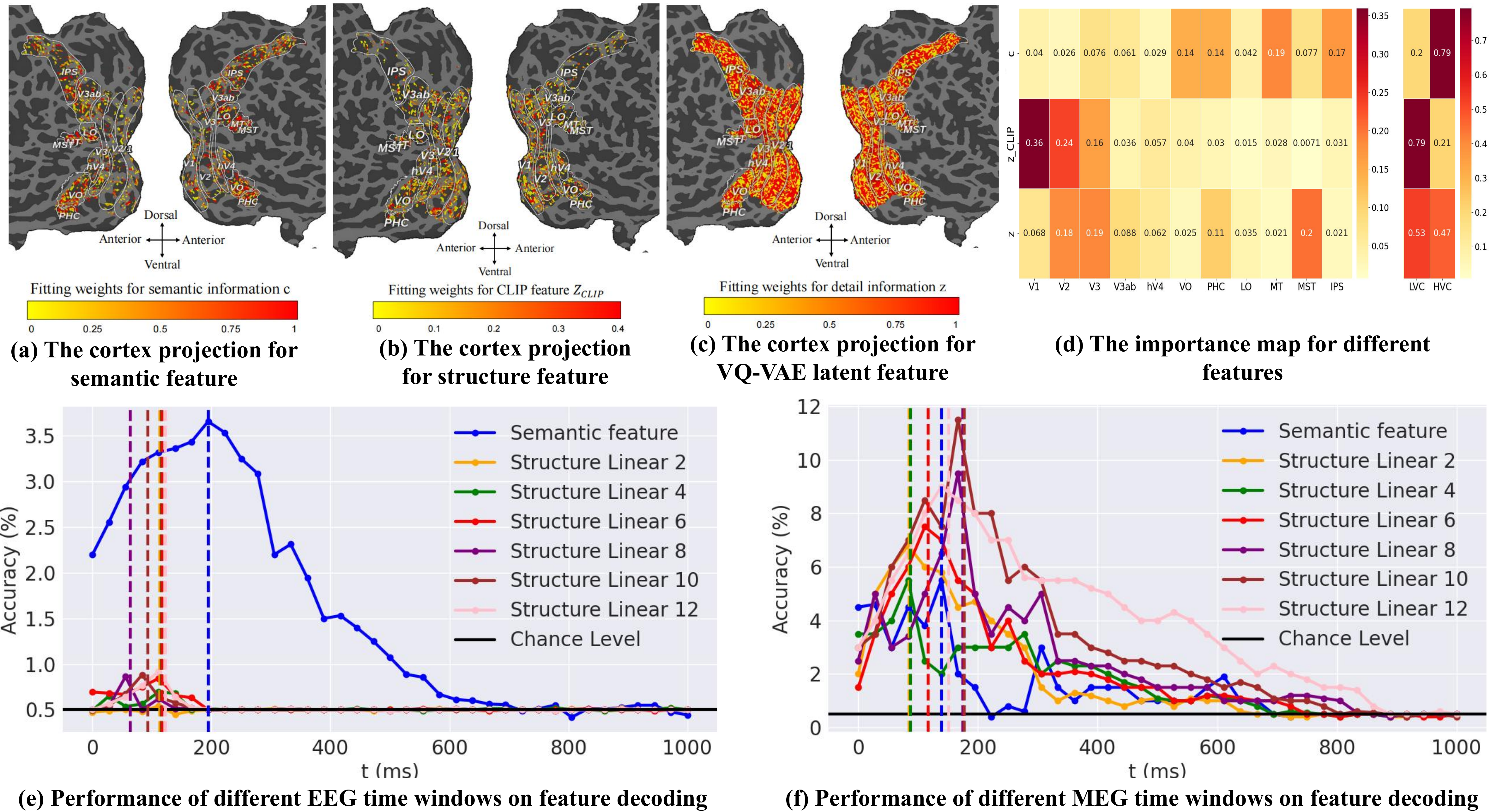}
	\caption{Spatial and temporal visualization results.  Subplots (a), (b), and (c) show the cortical projections for semantic features $c$, structural features $Z_{CLIP}$, and VQ-VAE latent space features $z$, respectively. Subplots (e) and (f) illustrate the changes in feature decoding accuracy using EEG and MEG data with different time windows, respectively.}
	\label{fig:visualcortexsub1}
\end{figure*}

\subsubsection{Spatial analysis} During the feature decoding process, we use linear regression models to fit three types of feature: semantic feature $c$, structural feature $Z_{CLIP}$, and VQ-VAE latent feature $z$. Once these linear models were trained, the weights at each voxel represent the importance of it in decoding a specific feature. We ultilize pycortex \cite{10.3389/fninf.2015.00023} to project the weights of each voxel in the trained model onto the visual cortex.


As shown in Fig. \ref{fig:visualcortexsub1}(a), when decoding the semantic feature \( c \), the voxels with the highest weights are predominantly located in brain regions such as VO, PHC, and MT, which are involved in processing high-level conceptual and semantic information. In contrast, Fig. \ref{fig:visualcortexsub1}(b) illustrates that when decoding the structural feature \( Z_{CLIP} \), the highest-weighted voxels are primarily concentrated in areas such as V1, V2, and V3, which are responsible for processing low-level texture and color information. Furthermore, Fig. \ref{fig:visualcortexsub1}(d) shows that, during the decoding of \( c \), HVC exhibits significantly greater weight than LVC, whereas the reverse holds for decoding \( Z_{CLIP} \). These findings are consistent with previous studies in neuroscience \cite{pinto2009high, krizhevsky2009learning,  Wang2022.09.27.508760}.

While previous works have used noisy VQ-VAE latent features as the starting point for denoising in diffusion models to achieve more precise reconstructions, they have not addressed the specific information encoded in these features or the brain regions responsible for interpreting them. This study addresses this question from a neuroscience perspective. As shown in Fig. \ref{fig:visualcortexsub1}(c), decoding the VQ-VAE latent feature \( z \) involves nearly all ROIs in the visual cortex. Additionally, Fig. \ref{fig:visualcortexsub1}(d) reveals that both LVC and HVC contribute similarly when decoding this feature, indicating that the VQ-VAE latent space likely represents intermediate features \cite{fu2023dreamsim} bridging high-level semantics and low-level structures, such as object contours and spatial layouts.
\subsubsection{Temporal analysis}
When decoding features using EEG or MEG data, we divided the data into fine-grained time windows with a 25 ms step size and a 100 ms window width. The change in Top-1 accuracy is shown in the curves in Figs. \ref{fig:visualcortexsub1}(e) and (f). As shown in this figure, peak decoding accuracy for EEG and MEG signals occurs around 100 to 200 ms after the subjects view the stimulus images, indicating that visual information is primarily processed during this period. Furthermore, comparing subfigures (e) and (f), we observe that although the peak decoding accuracy for structural features in EEG exceeds the chance level, it remains significantly lower than that of MEG. We attribute this to the lower spatial resolution of EEG, which captures less fine-grained structural information from brain activity. This also provides an explanation for the smaller improvement of our framework on EEG data compared to MEG, as presented in Table \ref{tab:result2}.

In summary, these spatial and temporal analyses offer tangible guidance for future neural decoding efforts. \textbf{Spatially}, our findings from fMRI data validate that researchers can target specific cortical regions to optimize decoding performance: the high-level visual cortex (HVC) is paramount for semantic content, while the low-level visual cortex (LVC) is crucial for structural details.  This provide a direct neuroscientific justification for designing decoding models with multi-pathway architectures, akin to our two-stage framework, that explicitly mirror the brain's functional separation of 'what' (semantic) and 'where' (structural) processing. \textbf{Temporally}, our analysis of EEG and MEG data pinpoints the 100-200 ms post-stimulus window as the most informative period for decoding visual information. Furthermore, the observed superiority of MEG over EEG in capturing fine-grained structural information suggests that the choice of imaging modality is critical; for reconstruction tasks demanding high structural fidelity, MEG should be the preferred modality, or advanced methods to enhance the spatial resolution of EEG should be considered.

\subsection{Limitations}
\subsubsection{Trade-off between structural alignment and semantic consistency}
Our framework employs a two-stage approach that decouples the injection of semantic and structural information into the reconstructed image. In Stage 1, we leverage decoded semantic features—specifically CLIP text embeddings—to generate an initial image that is semantically coherent. Subsequently, in Stage 2, this initial image is iteratively refined using decoded low-level features from shallow CLIP visual layers as a structural constraint. As demonstrated in Table 4, while this decoupled strategy yields substantial improvements across most evaluation metrics, it occasionally leads to a marginal decline in certain high-level semantic metrics. We attribute this trade-off to our sequential constraint strategy. Specifically, the optimization in Stage 2 is exclusively driven by structural objectives, forcing the model to prioritize the alignment of low-level information. This strong structural guidance, without a concurrent semantic preservation term, can inadvertently cause a slight ``semantic drift," thereby sacrificing a degree of high-level semantic consistency.

\subsubsection{Semantic-Regularized Structural Alignment}
To mitigate this trade-off, we introduce a simple yet effective semantic regularization term. Specifically, during the structural alignment in Stage 2, we augment the objective function. In addition to the structural loss ($\L_{\text{Structure}}$) which aligns shallow CLIP features, we incorporate a semantic constraint, denoted as $\L_{\text{Semantic}}$. This new term is derived from the features of the penultimate layer of the CLIP image encoder. The final objective function is a weighted sum controlled by a hyperparameter $\beta$, which governs the balance between semantic fidelity and structural alignment:
\begin{equation}
	\L_{\text{total}} = \L_{\text{Structure}} + \beta \cdot \L_{\text{Semantic}}  
\end{equation}

\begin{table}[htbp!]
	\caption{Impact of the semantic regularization hyperparameter ($\beta$) on reconstruction performance. We integrated our framework with the Takagi et al. \cite{takagi2023high} model on the fMRI dataset, and the results were averaged across four subjects. Green backgrounds indicate performance improvements, and red backgrounds indicate performance decreases.}
	\label{tab:result_addition}
	\vspace{8pt}
	\centering
	
		\scalebox{0.55}{ 
			\begin{tabular}{cccccccccccc}
				\toprule
				& \multicolumn{3}{c}{}                                                                                                        & \multicolumn{4}{c}{Low-level}                                                                                                           & \multicolumn{4}{c}{High-level}                                                                                                        \\ \cline{5-12} 
				\multirow{-2}{*}{Data type} & \multicolumn{3}{c}{\multirow{-2}{*}{Methods}}                                                                               & PixCorr$ \uparrow $             & SSIM$ \uparrow $                & AlexNet(2)$ \uparrow $          & AlexNet(5)$ \uparrow $          & InceptionV3$ \uparrow $         & CLIP$ \uparrow $                & EffNet-B$ \downarrow $            & SWAV$ \downarrow $                               \\ \hline
				
				&                                    &                                                          & $\beta = 0$         & \cellcolor[HTML]{C9E4B4}12.528\% & \cellcolor[HTML]{C9E4B4}7.154\%  & \cellcolor[HTML]{C9E4B4}4.313\%  & \cellcolor[HTML]{C9E4B4}1.416\% & \cellcolor[HTML]{F4B7BE}0.286\% & \cellcolor[HTML]{C9E4B4}0.701\% & \cellcolor[HTML]{C9E4B4}3.465\% & \cellcolor[HTML]{C9E4B4}4.669\% \\
				&                                    &                                                          & $\beta = 0.005$         & \cellcolor[HTML]{C9E4B4}12.463\% & \cellcolor[HTML]{C9E4B4}7.149\%  & \cellcolor[HTML]{C9E4B4}4.256\%  & \cellcolor[HTML]{C9E4B4}1.473\% & \cellcolor[HTML]{C9E4B4}1.458\% & \cellcolor[HTML]{C9E4B4}0.943\% & \cellcolor[HTML]{C9E4B4}3.677\% & \cellcolor[HTML]{C9E4B4}4.923\% \\
				&                                    &                                                          & $\beta = 0.05$         & \cellcolor[HTML]{C9E4B4}10.500\% & \cellcolor[HTML]{C9E4B4}6.567\%  & \cellcolor[HTML]{C9E4B4}2.153\%  & \cellcolor[HTML]{C9E4B4}0.916\% & \cellcolor[HTML]{C9E4B4}0.538\% & \cellcolor[HTML]{C9E4B4}0.764\% & \cellcolor[HTML]{C9E4B4}4.273\% & \cellcolor[HTML]{C9E4B4}4.732\% \\
				&                                    &                                                          & $\beta = 0.5$         & \cellcolor[HTML]{F4B7BE}6.745\% & \cellcolor[HTML]{F4B7BE}3.421\%  & \cellcolor[HTML]{F4B7BE}4.647\%  & \cellcolor[HTML]{F4B7BE}5.236\% & \cellcolor[HTML]{C9E4B4}1.723\% & \cellcolor[HTML]{C9E4B4}0.533\% & \cellcolor[HTML]{C9E4B4}4.256\% & \cellcolor[HTML]{C9E4B4}2.305\% \\
				\multirow{-5}{*}{fMRI}               &                                    & \multirow{-5}{*}{Takagi et al. \cite{takagi2023high}}        & $\beta = 1$        & \cellcolor[HTML]{F4B7BE}8.742\% & \cellcolor[HTML]{F4B7BE}8.653\%  & \cellcolor[HTML]{F4B7BE}4.562\%  & \cellcolor[HTML]{F4B7BE}3.328\% & \cellcolor[HTML]{C9E4B4}2.030\% & \cellcolor[HTML]{C9E4B4}1.075\% & \cellcolor[HTML]{C9E4B4}5.631\% & \cellcolor[HTML]{C9E4B4}3.109\% \\\hline
				&                                    &                                                          & $\beta = 0$         & \cellcolor[HTML]{C9E4B4}9.402\%          & \cellcolor[HTML]{C9E4B4}4.685\%  & \cellcolor[HTML]{C9E4B4}2.076\%  & \cellcolor[HTML]{C9E4B4}1.679\% & \cellcolor[HTML]{F4B7BE}0.730\% & \cellcolor[HTML]{C9E4B4}0.416\% & \cellcolor[HTML]{C9E4B4}0.623\% & \cellcolor[HTML]{C9E4B4}1.862\% \\
				\multirow{-2}{*}{EEG}               &                                    & \multirow{-2}{*}{ATM \cite{li2024visual}}        & $\beta = 0.005$        & \cellcolor[HTML]{C9E4B4}8.973\%          & \cellcolor[HTML]{C9E4B4}4.692\%  & \cellcolor[HTML]{C9E4B4}2.141\%  & \cellcolor[HTML]{C9E4B4}1.486\% & \cellcolor[HTML]{C9E4B4}0.942\% & \cellcolor[HTML]{C9E4B4}0.533\% & \cellcolor[HTML]{C9E4B4}0.769\% & \cellcolor[HTML]{C9E4B4}1.950\% \\\bottomrule
			\end{tabular}%
		}
\end{table}

We illustrate the impact of varying the hyperparameter $\beta$ on image reconstruction in Table~\ref{tab:result_addition}, with the experiment conducted on the Takagi et al.~\cite{takagi2023high} model using the fMRI dataset. The results confirm that applying a semantic regularization term during Stage 2 can successfully enhance low-level metrics while preventing the degradation of high-level ones. However, the outcome is highly sensitive to the choice of $\beta$. Optimal performance is achieved at a small value of $\beta = 0.005$. As $\beta$ is increased to 0.5 and further to 1.0, the semantic regularization term begins to dominate the optimization process, leading to a substantial decline in low-level metrics as the iterations proceed. Furthermore, we conducted experiments on the EEG dataset using the same hyperparameter settings, and the results confirmed the broad effectiveness of this regularization method.

\section{Conclusion}
This paper presents MindDiffuser, a two-stage image reconstruction framework that addresses the common limitations of previous methods by separately decoding semantic features and aligning structural features. We integrate this framework into existing SOTA reconstruction models and conduct extensive experiments on three large-scale neuroimaging datasets. The results demonstrate that our framework enhances the structural alignment of other models, validating its versatility. Finally, we employ spatial and temporal analyses to confirm the interpretability of our framework from a neuroscience perspective.

\section*{Acknowledgments}

This work was supported in part by the National Key R\&D Program of China (2023YFF1203501); in part by the National Natural Science Foundation of China under Grant 62576336, U2441253, 62206284 and 82272072;  in part by Beijing Natural Science Foundation under Grant L243016, and in part by the Strategic Priority Research Program of the Chinese Academy of Sciences (XDB0930000).

\section*{Declarations of Conflict of Interest}

The authors declared that they have no conflicts of interest to this work.

\bibliography{tmm}

@inproceedings{lu2023minddiffuser,
	title={Minddiffuser: Controlled image reconstruction from human brain activity with semantic and structural diffusion},
	author={Lu, Yizhuo and Du, Changde and Zhou, Qiongyi and Wang, Dianpeng and He, Huiguang},
	booktitle={Proceedings of the 31st ACM International Conference on Multimedia},
	pages={5899--5908},
	year={2023}
}

@article{zhou2025interpretable,
	title={Interpretable Visual Neural Decoding with Unsupervised Semantic Disentanglement},
	author={Zhou, Qiongyi and Du, Changde and Li, Dan and Wen, Bincheng and Chang, Le and He, Huiguang},
	journal={Machine Intelligence Research},
	pages={1--18},
	year={2025},
	publisher={Springer}
}

@article{zhou2022exploring,
	title={Exploring the brain-like properties of deep neural networks: a neural encoding perspective},
	author={Zhou, Qiongyi and Du, Changde and He, Huiguang},
	journal={Machine Intelligence Research},
	volume={19},
	number={5},
	pages={439--455},
	year={2022},
	publisher={Springer}
}

@ARTICLE{10047967,
	author={Huang, Zhongyu and Du, Changde and Wang, Yingheng and Fu, Kaicheng and He, Huiguang},
	journal={IEEE Transactions on Medical Imaging}, 
	title={Graph-Enhanced Emotion Neural Decoding}, 
	year={2023},
	volume={42},
	number={8},
	pages={2262-2273},
	doi={10.1109/TMI.2023.3246220}}

@article{benchetrit2023brain,
	title={Brain decoding: toward real-time reconstruction of visual perception},
	author={Benchetrit, Yohann and Banville, Hubert and King, Jean-R{\'e}mi},
	journal={International Conference on Learning Representations},
	year={2024}
}

@article{hebart2023things,
	title={THINGS-data, a multimodal collection of large-scale datasets for investigating object representations in human brain and behavior},
	author={Hebart, Martin N and Contier, Oliver and Teichmann, Lina and Rockter, Adam H and Zheng, Charles Y and Kidder, Alexis and Corriveau, Anna and Vaziri-Pashkam, Maryam and Baker, Chris I},
	journal={Elife},
	volume={12},
	pages={e82580},
	year={2023},
	publisher={eLife Sciences Publications Limited}
}

@ARTICLE{10858771,
	author={Ma, Yongqiang and Liu, Yulong and Chen, Liangjun and Zhu, Guibo and Chen, Badong and Zheng, Nanning},
	journal={IEEE Transactions on Medical Imaging}, 
	title={BrainCLIP: Brain Representation via CLIP for Generic Natural Visual Stimulus Decoding}, 
	year={2025},
	volume={},
	number={},
	pages={1-1},
	doi={10.1109/TMI.2025.3537287}}

@article{vaziri2017goal,
	title={Goal-directed visual processing differentially impacts human ventral and dorsal visual representations},
	author={Vaziri-Pashkam, Maryam and Xu, Yaoda},
	journal={Journal of Neuroscience},
	volume={37},
	number={36},
	pages={8767--8782},
	year={2017},
	publisher={Soc Neuroscience}
}

@article{donahue2019large,
	title={Large scale adversarial representation learning},
	author={Donahue, Jeff and Simonyan, Karen},
	journal={Advances in neural information processing systems},
	volume={32},
	year={2019}
}

@article{zachariou2014ventral,
	title={Ventral and dorsal visual stream contributions to the perception of object shape and object location},
	author={Zachariou, Valentinos and Klatzky, Roberta and Behrmann, Marlene},
	journal={Journal of Cognitive Neuroscience},
	volume={26},
	number={1},
	pages={189--209},
	year={2014},
	publisher={MIT Press One Rogers Street, Cambridge, MA 02142-1209, USA journals-info~…}
}

@ARTICLE{10.3389/fncom.2019.00021,
	
	AUTHOR={Shen, Guohua and Dwivedi, Kshitij and Majima, Kei and Horikawa, Tomoyasu and Kamitani, Yukiyasu},   
	
	TITLE={End-to-End Deep Image Reconstruction From Human Brain Activity},      
	
	JOURNAL={Frontiers in Computational Neuroscience},      
	
	VOLUME={13},           
	
	YEAR={2019},      
	
	URL={https://www.frontiersin.org/articles/10.3389/fncom.2019.00021},       
	
	DOI={10.3389/fncom.2019.00021},      
	
	ISSN={1662-5188}
}

@article{rakhimberdina2021natural,
	title={Natural image reconstruction from f{MRI} using deep learning: A survey},
	author={Rakhimberdina, Zarina and Jodelet, Quentin and Liu, Xin and Murata, Tsuyoshi},
	journal={Frontiers in neuroscience},
	volume={15},
	pages={795488},
	year={2021},
	publisher={Frontiers Media SA}
}

@inproceedings{xia2024dream,
	title={Dream: Visual decoding from reversing human visual system},
	author={Xia, Weihao and de Charette, Raoul and Oztireli, Cengiz and Xue, Jing-Hao},
	booktitle={Proceedings of the IEEE/CVF Winter Conference on Applications of Computer Vision},
	pages={8226--8235},
	year={2024}
}

@article{kneeland2023second,
	title={Second Sight: Using brain-optimized encoding models to align image distributions with human brain activity},
	author={Kneeland, Reese and Ojeda, Jordyn and St-Yves, Ghislain and Naselaris, Thomas},
	journal={ArXiv},
	year={2023},
	publisher={ArXiv}
}

@inproceedings{ozcelik2022reconstruction,
	title={Reconstruction of perceived images from fmri patterns and semantic brain exploration using instance-conditioned gans},
	author={Ozcelik, Furkan and Choksi, Bhavin and Mozafari, Milad and Reddy, Leila and VanRullen, Rufin},
	booktitle={2022 International Joint Conference on Neural Networks (IJCNN)},
	pages={1--8},
	year={2022},
	organization={IEEE}
}

@article{du2020structured,
	title={Structured neural decoding with multitask transfer learning of deep neural network representations},
	author={Du, Changde and Du, Changying and Huang, Lijie and Wang, Haibao and He, Huiguang},
	journal={IEEE Transactions on Neural Networks and Learning Systems},
	volume={33},
	number={2},
	pages={600--614},
	year={2020},
	publisher={IEEE}
}

@article{fujiwara2013modular,
	title={Modular encoding and decoding models derived from {B}ayesian canonical correlation analysis},
	author={Fujiwara, Yusuke and Miyawaki, Yoichi and Kamitani, Yukiyasu},
	journal={Neural computation},
	volume={25},
	number={4},
	pages={979--1005},
	year={2013},
	publisher={MIT Press One Rogers Street, Cambridge, MA 02142-1209, USA journals-info~…}
}

@inproceedings{xu2023versatile,
	title={Versatile diffusion: Text, images and variations all in one diffusion model},
	author={Xu, Xingqian and Wang, Zhangyang and Zhang, Gong and Wang, Kai and Shi, Humphrey},
	booktitle={Proceedings of the IEEE/CVF International Conference on Computer Vision},
	pages={7754--7765},
	year={2023}
}

@article{fu2023dreamsim,
	title={Dreamsim: Learning new dimensions of human visual similarity using synthetic data},
	author={Fu, Stephanie and Tamir, Netanel and Sundaram, Shobhita and Chai, Lucy and Zhang, Richard and Dekel, Tali and Isola, Phillip},
	journal={Advances in neural information processing systems},
	year={2023}
}

@article{pinto2009high,
	title={A high-throughput screening approach to discovering good forms of biologically inspired visual representation},
	author={Pinto, Nicolas and Doukhan, David and DiCarlo, James J and Cox, David D},
	journal={PLoS computational biology},
	volume={5},
	number={11},
	pages={e1000579},
	year={2009},
	publisher={Public Library of Science San Francisco, USA}
}

@inproceedings{xie2024brainram,
	title={BrainRAM: Cross-Modality Retrieval-Augmented Image Reconstruction from Human Brain Activity},
	author={Xie, Dian and Zhao, Peiang and Zhang, Jiarui and Wei, Kangqi and Ni, Xiaobao and Xia, Jiong},
	booktitle={Proceedings of the 32nd ACM International Conference on Multimedia},
	pages={3994--4003},
	year={2024}
}

@article{ozcelik2023natural,
	title={Natural scene reconstruction from fMRI signals using generative latent diffusion},
	author={Ozcelik, Furkan and VanRullen, Rufin},
	journal={Scientific Reports},
	volume={13},
	number={1},
	pages={15666},
	year={2023},
	publisher={Nature Publishing Group UK London}
}

@article{song2023decoding,
	title={Decoding Natural Images from EEG for Object Recognition},
	author={Song, Yonghao and Liu, Bingchuan and Li, Xiang and Shi, Nanlin and Wang, Yijun and Gao, Xiaorong},
	journal={International Conference on Learning Representations},
	year={2024}
}

@article{ye2023ip,
	title={Ip-adapter: Text compatible image prompt adapter for text-to-image diffusion models},
	author={Ye, Hu and Zhang, Jun and Liu, Sibo and Han, Xiao and Yang, Wei},
	journal={arXiv preprint arXiv:2308.06721},
	year={2023}
}

@article{lawhern2018eegnet,
	title={EEGNet: a compact convolutional neural network for EEG-based brain--computer interfaces},
	author={Lawhern, Vernon J and Solon, Amelia J and Waytowich, Nicholas R and Gordon, Stephen M and Hung, Chou P and Lance, Brent J},
	journal={Journal of neural engineering},
	volume={15},
	number={5},
	pages={056013},
	year={2018},
	publisher={iOP Publishing}
}

@article{krizhevsky2012imagenet,
	title={Imagenet classification with deep convolutional neural networks},
	author={Krizhevsky, Alex and Sutskever, Ilya and Hinton, Geoffrey E},
	journal={Advances in neural information processing systems},
	volume={25},
	year={2012}
}

@inproceedings{szegedy2016rethinking,
	title={Rethinking the inception architecture for computer vision},
	author={Szegedy, Christian and Vanhoucke, Vincent and Ioffe, Sergey and Shlens, Jon and Wojna, Zbigniew},
	booktitle={Proceedings of the IEEE conference on computer vision and pattern recognition},
	pages={2818--2826},
	year={2016}
}

@inproceedings{radford2021learning,
	title={Learning transferable visual models from natural language supervision},
	author={Radford, Alec and Kim, Jong Wook and Hallacy, Chris and Ramesh, Aditya and Goh, Gabriel and Agarwal, Sandhini and Sastry, Girish and Askell, Amanda and Mishkin, Pamela and Clark, Jack and others},
	booktitle={International conference on machine learning},
	pages={8748--8763},
	year={2021},
	organization={PMLR}
}

@article{caron2020unsupervised,
	title={Unsupervised learning of visual features by contrasting cluster assignments},
	author={Caron, Mathilde and Misra, Ishan and Mairal, Julien and Goyal, Priya and Bojanowski, Piotr and Joulin, Armand},
	journal={Advances in neural information processing systems},
	volume={33},
	pages={9912--9924},
	year={2020}
}

@inproceedings{tan2019efficientnet,
	title={Efficientnet: Rethinking model scaling for convolutional neural networks},
	author={Tan, Mingxing and Le, Quoc},
	booktitle={International conference on machine learning},
	pages={6105--6114},
	year={2019},
	organization={PMLR}
}

@article{wang2004image,
	title={Image quality assessment: from error visibility to structural similarity},
	author={Wang, Zhou and Bovik, Alan C and Sheikh, Hamid R and Simoncelli, Eero P},
	journal={IEEE transactions on image processing},
	volume={13},
	number={4},
	pages={600--612},
	year={2004},
	publisher={IEEE}
}

@article{gifford2022large,
	title={A large and rich EEG dataset for modeling human visual object recognition},
	author={Gifford, Alessandro T and Dwivedi, Kshitij and Roig, Gemma and Cichy, Radoslaw M},
	journal={NeuroImage},
	volume={264},
	pages={119754},
	year={2022},
	publisher={Elsevier}
}

@article{scotti2024reconstructing,
	title={Reconstructing the {M}ind's {E}ye: f{MRI}-to-image with contrastive learning and diffusion priors},
	author={Scotti, Paul and Banerjee, Atmadeep and Goode, Jimmie and Shabalin, Stepan and Nguyen, Alex and Dempster, Aidan and Verlinde, Nathalie and Yundler, Elad and Weisberg, David and Norman, Kenneth and others},
	journal={Advances in Neural Information Processing Systems},
	volume={36},
	year={2024}
}

@article{naselaris2009bayesian,
	title={Bayesian reconstruction of natural images from human brain activity},
	author={Naselaris, Thomas and Prenger, Ryan J and Kay, Kendrick N and Oliver, Michael and Gallant, Jack L},
	journal={Neuron},
	volume={63},
	number={6},
	pages={902--915},
	year={2009},
	publisher={Elsevier}
}

@article{simonyan2014very,
	title={Very deep convolutional networks for large-scale image recognition},
	author={Simonyan, Karen},
	journal={arXiv preprint arXiv:1409.1556},
	year={2014}
}

@article{shen2019deep,
	title={Deep image reconstruction from human brain activity},
	author={Shen, Guohua and Horikawa, Tomoyasu and Majima, Kei and Kamitani, Yukiyasu},
	journal={PLoS computational biology},
	volume={15},
	number={1},
	pages={e1006633},
	year={2019},
	publisher={Public Library of Science San Francisco, CA USA}
}

@article{lin2022mind,
	title={Mind {R}eader: Reconstructing {C}omplex {I}mages from {B}rain {A}ctivities},
	author={Lin, Sikun and Sprague, Thomas and Singh, Ambuj K},
	journal={Advances in Neural Information Processing Systems},
	volume={35},
	pages={29624--29636},
	year={2022}
}

@article{du2018reconstructing,
	title={Reconstructing perceived images from human brain activities with {B}ayesian deep multiview learning},
	author={Du, Changde and Du, Changying and Huang, Lijie and He, Huiguang},
	journal={IEEE transactions on neural networks and learning systems},
	volume={30},
	number={8},
	pages={2310--2323},
	year={2018},
	publisher={IEEE}
}

@article{beliy2019voxels,
	title={From voxels to pixels and back: {S}elf-supervision in natural-image reconstruction from f{MRI}},
	author={Beliy, Roman and Gaziv, Guy and Hoogi, Assaf and Strappini, Francesca and Golan, Tal and Irani, Michal},
	journal={Advances in Neural Information Processing Systems},
	volume={32},
	year={2019}
}

@article{li2024visual,
	title={Visual decoding and reconstruction via eeg embeddings with guided diffusion},
	author={Li, Dongyang and Wei, Chen and Li, Shiying and Zou, Jiachen and Qin, Haoyang and Liu, Quanying},
	journal={Advances in Neural Information Processing Systems},
	year={2024}
}

@article{casanova2021instance,
	title={Instance-{C}onditioned {GAN}},
	author={Casanova, Arantxa and Careil, Marlene and Verbeek, Jakob and Drozdzal, Michal and Romero Soriano, Adriana},
	journal={Advances in Neural Information Processing Systems},
	volume={34},
	pages={27517--27529},
	year={2021}
}

@inproceedings{takagi2023high,
	title={High-resolution image reconstruction with latent diffusion models from human brain activity},
	author={Takagi, Yu and Nishimoto, Shinji},
	booktitle={Proceedings of the IEEE/CVF Conference on Computer Vision and Pattern Recognition},
	pages={14453--14463},
	year={2023}
}

@article{gu2022decoding,
	title={Decoding natural image stimuli from fmri data with a surface-based convolutional network},
	author={Gu, Zijin and Jamison, Keith and Kuceyeski, Amy and Sabuncu, Mert},
	journal={arXiv preprint arXiv:2212.02409},
	year={2022}
}

@article{gaziv2022self,
	title={Self-supervised natural image reconstruction and large-scale semantic classification from brain activity},
	author={Gaziv, Guy and Beliy, Roman and Granot, Niv and Hoogi, Assaf and Strappini, Francesca and Golan, Tal and Irani, Michal},
	journal={NeuroImage},
	volume={254},
	pages={119121},
	year={2022},
	publisher={Elsevier}
}

@article{wijmans1995solution,
	title={The solution-diffusion model: a review},
	author={Wijmans, Johannes G and Baker, Richard W},
	journal={Journal of membrane science},
	volume={107},
	number={1-2},
	pages={1--21},
	year={1995},
	publisher={Elsevier}
}

@article{ramesh2022hierarchical,
	title={Hierarchical text-conditional image generation with clip latents},
	author={Ramesh, Aditya and Dhariwal, Prafulla and Nichol, Alex and Chu, Casey and Chen, Mark},
	journal={arXiv preprint arXiv:2204.06125},
	volume={1},
	number={2},
	pages={3},
	year={2022}
}

@inproceedings{rombach2022high,
	title={High-resolution image synthesis with latent diffusion models},
	author={Rombach, Robin and Blattmann, Andreas and Lorenz, Dominik and Esser, Patrick and Ommer, Bj{\"o}rn},
	booktitle={Proceedings of the IEEE/CVF Conference on Computer Vision and Pattern Recognition},
	pages={10684--10695},
	year={2022}
}

@inproceedings{mou2024t2i,
	title={T2i-adapter: Learning adapters to dig out more controllable ability for text-to-image diffusion models},
	author={Mou, Chong and Wang, Xintao and Xie, Liangbin and Wu, Yanze and Zhang, Jian and Qi, Zhongang and Shan, Ying},
	booktitle={Proceedings of the AAAI Conference on Artificial Intelligence},
	volume={38},
	number={5},
	pages={4296--4304},
	year={2024}
}

@inproceedings{ronneberger2015u,
	title={U-net: {C}onvolutional networks for biomedical image segmentation},
	author={Ronneberger, Olaf and Fischer, Philipp and Brox, Thomas},
	booktitle={Medical Image Computing and Computer-Assisted Intervention--MICCAI 2015: 18th International Conference, Munich, Germany, October 5-9, 2015, Proceedings, Part III 18},
	pages={234--241},
	year={2015},
	organization={Springer}
}

@article{allen2022massive,
	title={A massive 7{T} f{MRI} dataset to bridge cognitive neuroscience and artificial intelligence},
	author={Allen, Emily J and St-Yves, Ghislain and Wu, Yihan and Breedlove, Jesse L and Prince, Jacob S and Dowdle, Logan T and Nau, Matthias and Caron, Brad and Pestilli, Franco and Charest, Ian and others},
	journal={Nature neuroscience},
	volume={25},
	number={1},
	pages={116--126},
	year={2022},
	publisher={Nature Publishing Group US New York}
}

@article {Wang2022.09.27.508760,
	author = {Aria Y. Wang and Kendrick Kay and Thomas Naselaris and Michael J. Tarr and Leila Wehbe},
	title = {Better models of human high-level visual cortex emerge from natural language supervision with a large and diverse dataset.},
	journal = {Nat Mach Intell 5, 1415–1426 (2023). https://doi.org/10.1038/s42256-023-00753-y}
}

@ARTICLE{10.3389/fninf.2015.00023,
	
	AUTHOR={Gao, James S. and Huth, Alexander G. and Lescroart, Mark D. and Gallant, Jack L.},   
	
	TITLE={Pycortex: an interactive surface visualizer for f{MRI}},      
	
	JOURNAL={Frontiers in Neuroinformatics},      
	
	VOLUME={9},           
	
	YEAR={2015},      
	
	URL={https://www.frontiersin.org/articles/10.3389/fninf.2015.00023},       
	
	DOI={10.3389/fninf.2015.00023},      
	
	ISSN={1662-5196}
}

@article{krizhevsky2009learning,
	title={Learning multiple layers of features from tiny images},
	author={Krizhevsky, Alex and Hinton, Geoffrey and others},
	year={2009},
	publisher={Toronto, ON, Canada}
}

@inproceedings{lin2014microsoft,
	title={Microsoft {COCO}: Common objects in context},
	author={Lin, Tsung-Yi and Maire, Michael and Belongie, Serge and Hays, James and Perona, Pietro and Ramanan, Deva and Doll{\'a}r, Piotr and Zitnick, C Lawrence},
	booktitle={Computer Vision--ECCV 2014: 13th European Conference, Zurich, Switzerland, September 6-12, 2014, Proceedings, Part V 13},
	pages={740--755},
	year={2014},
	organization={Springer}
}

@article{kingma2013auto,
	title={Auto-{E}ncoding {V}ariational {B}ayes},
	author={Kingma, Diederik P and Welling, Max},
	journal={International Conference on Learning Representations},
	year={2014}
}

@article{goodfellow2020generative,
	title={Generative {A}dversarial {N}etworks},
	author={Goodfellow, Ian and Pouget-Abadie, Jean and Mirza, Mehdi and Xu, Bing and Warde-Farley, David and Ozair, Sherjil and Courville, Aaron and Bengio, Yoshua},
	journal={Communications of the ACM},
	volume={63},
	number={11},
	pages={139--144},
	year={2020},
	publisher={ACM New York, NY, USA}
}

@article{ho2020denoising,
	title={Denoising {D}iffusion {P}robabilistic {M}odels},
	author={Ho, Jonathan and Jain, Ajay and Abbeel, Pieter},
	journal={Advances in neural information processing systems},
	volume={33},
	pages={6840--6851},
	year={2020}
}

@article{kay2008naselaris,
	title={Naselaris {T}, {P}renger {RJ}, {G}allant {JL}},
	author={Kay, KN},
	journal={Identifying natural images from human brain activity. nature},
	volume={452},
	pages={352--355},
	year={2008}
}

@article{chapelle2009semi,
	title={Semi-supervised learning (chapelle, o. et al., eds.; 2006)[book reviews]},
	author={Chapelle, Olivier and Scholkopf, Bernhard and Zien, Alexander},
	journal={IEEE Transactions on Neural Networks},
	volume={20},
	number={3},
	pages={542--542},
	year={2009},
	publisher={IEEE}
}

@article{chen2022seeing,
	title={Seeing {B}eyond the {B}rain: {C}onditional {D}iffusion {M}odel with {S}parse {M}asked {M}odeling for {V}ision {D}ecoding},
	author={Chen, Zijiao and Qing, Jiaxin and Xiang, Tiange and Yue, Wan Lin and Zhou, Juan Helen},
	journal={arXiv preprint arXiv:2211.06956},
	year={2022}
}

@article{buzsaki2012origin,
	title={The origin of extracellular fields and currents—EEG, ECoG, LFP and spikes},
	author={Buzs{\'a}ki, Gy{\"o}rgy and Anastassiou, Costas A and Koch, Christof},
	journal={Nature reviews neuroscience},
	volume={13},
	number={6},
	pages={407--420},
	year={2012},
	publisher={Nature Publishing Group UK London}
}

@article{henson2009selecting,
	title={Selecting forward models for MEG source-reconstruction using model-evidence},
	author={Henson, Richard N and Mattout, J{\'e}r{\'e}mie and Phillips, Christophe and Friston, Karl J},
	journal={Neuroimage},
	volume={46},
	number={1},
	pages={168--176},
	year={2009},
	publisher={Elsevier}
}

@inproceedings{quan2024psychometry,
	title={Psychometry: An omnifit model for image reconstruction from human brain activity},
	author={Quan, Ruijie and Wang, Wenguan and Tian, Zhibo and Ma, Fan and Yang, Yi},
	booktitle={Proceedings of the IEEE/CVF Conference on Computer Vision and Pattern Recognition},
	pages={233--243},
	year={2024}
}

@inproceedings{xia2024umbrae,
	title={Umbrae: Unified multimodal brain decoding},
	author={Xia, Weihao and de Charette, Raoul and Oztireli, Cengiz and Xue, Jing-Hao},
	booktitle={European Conference on Computer Vision},
	pages={242--259},
	year={2024},
	organization={Springer}
}

\begin{figure}[h]%
\centering
\includegraphics[width=0.3\textwidth]{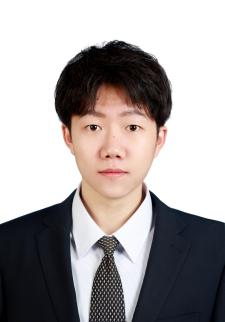}
\end{figure}

\noindent{\bf Yizhuo Lu}\quad received the B.S. degree in statistics from Beijing Institute of Technology, Beijing, China,
in 2023. He is a Ph.D. degree candidate under supervision of Dr. Huiguang He from
the Institute of Automation, Chinese Academy of Sciences, China. He has published papers at ICLR and ACM MM. His homepage is: \url{https://reedonepeck.github.io/Luyizhuo.github.io/}.

His research interests include deep learning, neural encoding and decoding, crossmodal generation, and brain-inspired intelligence.

E-mail: luyizhuo2023@ia.ac.cn

ORCID iD: 0009-0002-1196-6372

\newpage


\begin{figure}[h]%
\centering
\includegraphics[width=0.3\textwidth]{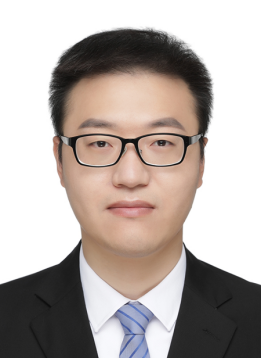}
\end{figure}

\noindent{\bf Changde Du }\quad received the Ph.D. degree
in technology of computer application
from the Institute of Automation, Chinese
Academy of Sciences, China in 2019. He
has published over 50 peer-reviewed research papers in prestigious conferences
and journals. He won the following awards:
National Scholarship for Doctoral Students (2018), President Prize of Chinese
Academy of Sciences for Excellent Ph.D. Graduates (2019). He is
currently an associate professor at Institute of Automation,
Chinese Academy of Sciences, China. His homepage is: \url{https://
	changdedu.github.io/}.

His research interests include deep learning, computational
neuroscience, brain-inspired intelligence, computer vision and
brain-computer interfaces.

E-mail: changde.du@ia.ac.cn


\begin{figure}[h]%
\centering
\includegraphics[width=0.3\textwidth]{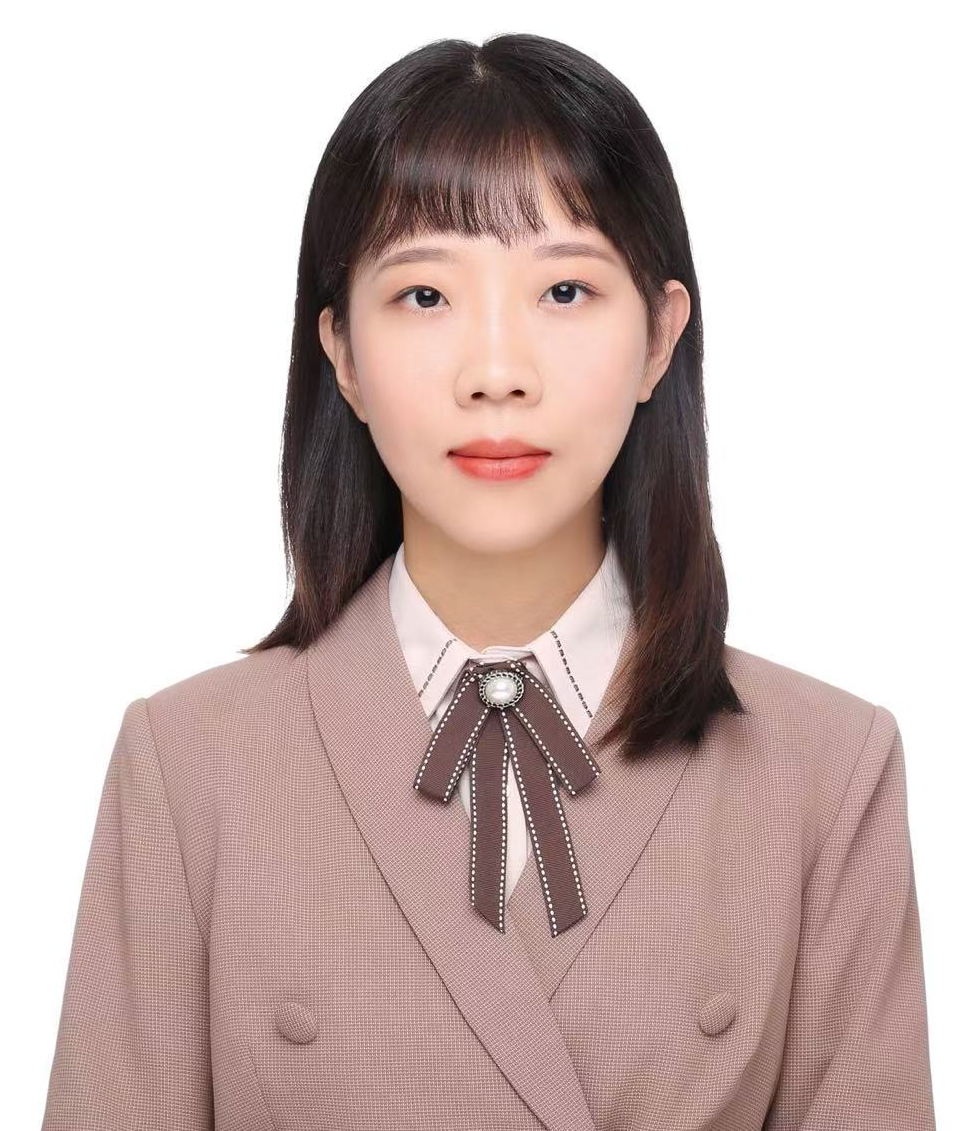}
\end{figure}

\noindent{\bf Qiongyi Zhou }\quad received the Ph.D. degree
in pattern recognition and intelligence systems from the Institute of Automation,
Chinese Academy of Sciences, China in
2023. She is currently a senior AI algorithm engineer at Honor Device Co., Ltd,
China.

Her research interests include deep learning, neural encoding and decoding, crossmodal generation, and brain-inspired intelligence.

E-mail: zhouqiongyi@hotmail.com

\newpage

\begin{figure}[h]%
	\centering
	\includegraphics[width=0.3\textwidth]{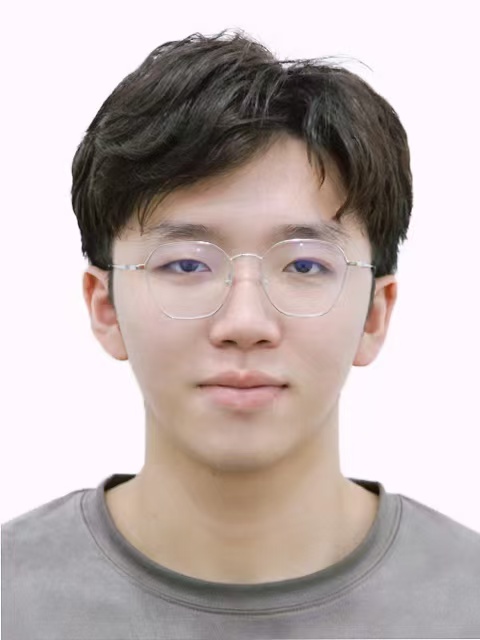}
\end{figure}

\noindent{\bf Liuyun Jiang}\quad received the B.S. degree in Automation from Beijing Institute of Technology, Beijing, China, in 2023. He is currently pursuing the Ph.D. degree under the supervision of Prof. Hua Han at the Institute of Automation, Chinese Academy of Sciences, Beijing, China. 

His research interests include computer vision, volume electron microscopy, and connectomics.

E-mail: jiangliuyun2023@ia.ac.cn

\begin{figure}[h]%
	\centering
	\includegraphics[width=0.3\textwidth]{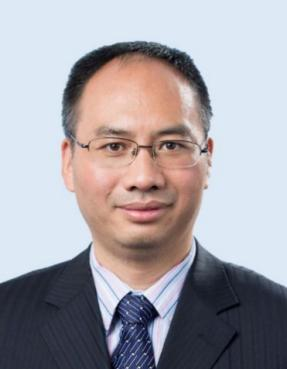}
\end{figure}

\noindent{\bf Huiguang He }\quad received the B.Sc. degree in
maritime traffic administration and the
M.Sc. degree in maritime traffic engineering from Dalian Maritime University
(DMU) China, in 1994 and 1997, respectively, and the Ph.D. degree (with honor) in
pattern recognition and intelligent systems from Institute of Automation,
Chinese Academy of Sciences (CASIA),
China in 2002. He was an associate lecturer in DMU from 1997 to
1999, and postdoctoral researcher in University of Rochester,
USA from 2003 to 2004. He was a visiting professor in University of North Carolina at Chapel Hill, USA from 2014 to 2015.
He is currently a full professor with CASIA. His research has
been supported by several research grants from National Science
Foundation of China, and he has published more than 180 peerreviewed papers. He won the following awards: Excellent Ph.D.
dissertation of Chinese Academy of Sciences (2004), National
Science \& Technology Award (2003, 2004), Beijing Science \&
Technology Award (2002, 2003), K.C. Wong Education Prizes
(2007, 2009), Jia-Xi Lu Young Talent Prize (2009) and excellent
member of Youth Innovation Promotion Association, CAS
(2016). He is a senior member of the IEEE.

His research interests include pattern recognition, medical image processing, and brain computer interface (BCI).

E-mail: huiguang.he@ia.ac.cn (Corresponding author)

ORCID iD: 0000-0002-0684-1711

\end{document}